\documentclass[10pt,twocolumn,letterpaper]{article}

\usepackage[pagenumbers]{cvpr} %

\usepackage{graphicx}
\usepackage{amsmath}
\usepackage{amssymb}
\usepackage{booktabs}
\usepackage{pifont}
\usepackage{float}
\usepackage[font=small,labelfont=bf,tableposition=top]{caption}
\usepackage{multirow}
\usepackage{bm}
\usepackage{enumitem}

\newcommand{\e}[1]{{\small $#1$}}

\newcounter{alphasect}
\def\alphainsection{0}

\let\oldsection=\section
\def\section{%
  \ifnum\alphainsection=1%
    \addtocounter{alphasect}{1}
  \fi%
\oldsection}%

\renewcommand\thesection{%
  \ifnum\alphainsection=1%
    \Alph{alphasect}%
  \else
    \arabic{section}%
  \fi%
}%

\newenvironment{alphasection}{%
  \ifnum\alphainsection=1%
    \errhelp={Let other blocks end at the beginning of the next block.}
    \errmessage{Nested Alpha section not allowed}
  \fi%
  \setcounter{alphasect}{0}
  \def\alphainsection{1}
}{%
  \setcounter{alphasect}{0}
  \def\alphainsection{0}
}%

\usepackage[pagebackref,breaklinks,colorlinks]{hyperref}

\usepackage[capitalize]{cleveref}
\crefname{section}{Sec.}{Secs.}
\Crefname{section}{Section}{Sections}
\Crefname{table}{Table}{Tables}
\crefname{table}{Tab.}{Tabs.}

\begin{document}

\title{Uncovering the Disentanglement Capability in Text-to-Image Diffusion Models}

\author{
\textbf{Qiucheng Wu}${^1}$, \textbf{Yujian Liu}$^{1}$,
\textbf{Handong Zhao}${^2}$,\\ \textbf{Ajinkya Kale}${^2}$, \textbf{Trung Bui}${^2}$, \textbf{Tong Yu}${^2}$, \textbf{Zhe Lin}${^2}$,\\ \textbf{Yang Zhang}${^3}$, \textbf{Shiyu Chang}${^1}$\\
$^1$UC, Santa Barbara, $^2$Adobe Research, $^3$ MIT-IBM Watson AI Lab
\\\small\texttt{\{qiucheng, yujianliu\}@ucsb.edu}}

\let\oldtwocolumn\twocolumn
\renewcommand\twocolumn[1][]{%
    \oldtwocolumn[{#1}{
    \begin{center}

\includegraphics[width=\textwidth]{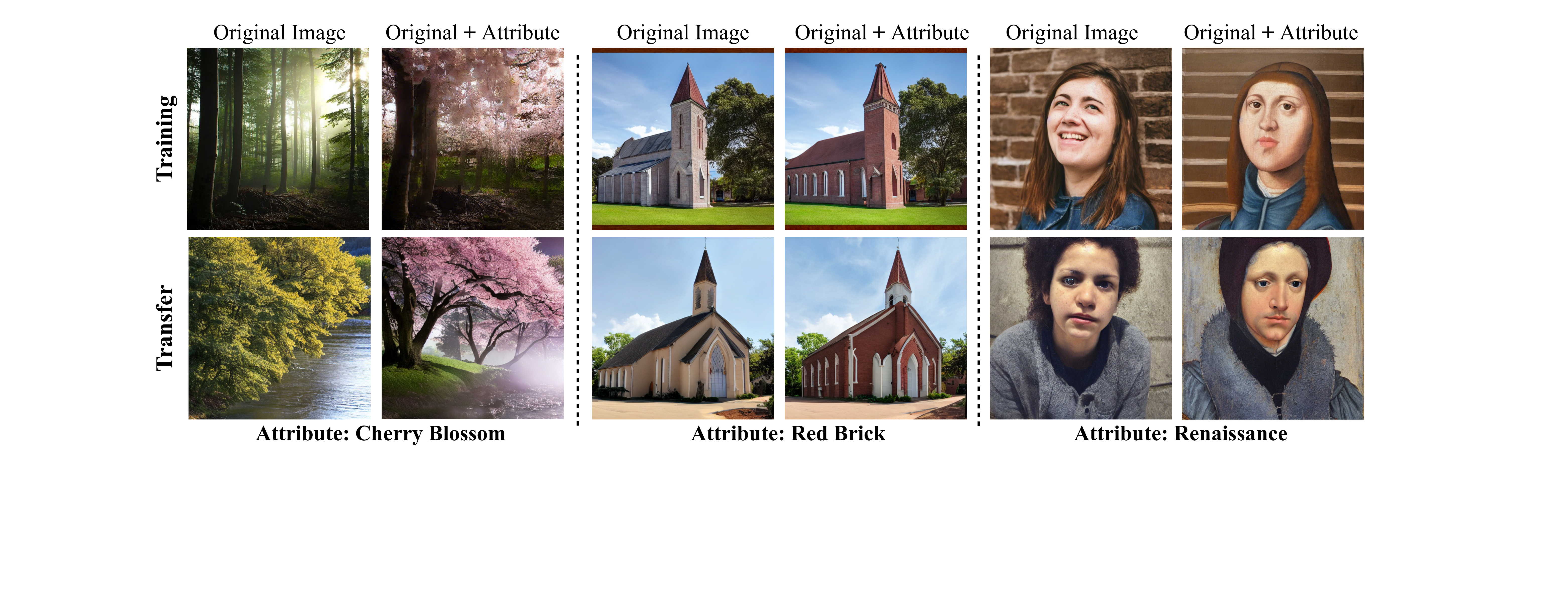}
    \captionof{figure}{\textbf{Example attributes disentangled from the stable diffusion model.}
    Based on a \textit{fixed} stable diffusion model, we disentangle the target attribute from a single training image. The learned parameters can then be applied to an unseen image and achieve the same edit.}
    \label{fig:intro}
        \end{center}
    }]
}

\maketitle

\begin{abstract}
Generative models have been widely studied in computer vision. Recently, diffusion models have drawn substantial attention due to the high quality of their generated images. A key desired property of image generative models is the ability to disentangle different attributes, which should enable modification towards a style without changing the semantic content, and the modification parameters should generalize to different images. Previous studies have found that generative adversarial networks (GANs) are inherently endowed with such disentanglement capability, so they can perform disentangled image editing without re-training or fine-tuning the network. In this work, we explore whether diffusion models are also inherently equipped with such a capability. Our finding is that for stable diffusion models, by partially changing the input text embedding from a neutral description (\emph{e.g.,} ``a photo of person'') to one with style (\emph{e.g.,} ``a photo of person with smile'') while fixing all the Gaussian random noises introduced during the denoising process, the generated images can be modified towards the target style without changing the semantic content. Based on this finding, we further propose a simple, light-weight image editing algorithm where the mixing weights of the two text embeddings are optimized for style matching and content preservation. This entire process only involves optimizing over around 50 parameters and does not fine-tune the diffusion model itself. Experiments show that the proposed method can modify a wide range of attributes, with the performance outperforming diffusion-model-based image-editing algorithms that require fine-tuning. The optimized weights generalize well to different images.  Our code is publicly available at \url{https://github.com/UCSB-NLP-Chang/DiffusionDisentanglement}.
\end{abstract}

\vspace{-1.8em}
\section{Introduction}
\label{sec:intro}

\begin{table*}[t]
\centering
\begin{tabular}{clll}
\toprule
  &   & \textbf{Scenes} & \textbf{Person} \\ 
\midrule 
\multirow{3}{*}{\textcolor{green}{\ding{52}}}  & \multirow{2}{*}{\textbf{Global}} &
{\footnotesize \textbf{Styles} (children drawing, cyberpunk, anime), \textbf{Building appearance}} & {\footnotesize \textbf{Styles} (renaissance, Egyptian mural, sketch, Pixar)} \\
& & {\footnotesize (wooden, red brick), \textbf{Weather \& time} (sunset, night, snowy) } & {\footnotesize \textbf{Appearance} (young, tanned, male) } \\
\cmidrule(lr){2-4}
& \textbf{Local} & {\footnotesize Cherry blossom, rainbow, foothills}  &  {\footnotesize \textbf{Expressions} (smiling, crying, angry)} \\
\midrule
\textcolor{red}{\ding{56}} & \textbf{Small edits}
& {\footnotesize Cake toppings, remove people on the street} &
{\footnotesize Hats, hair colors, earrings}  \\
\bottomrule

\end{tabular}
\caption{\textbf{Summarization of explored attributes.}
\textcolor{green}{\ding{52}} shows successfully disentangled attributes and \textcolor{red}{\ding{56}} shows failure cases.
Small edits on the image are harder to be disentangled when the target attribute correlates with other parts of the image.
}
\vspace*{-0.2in}
\label{tab:attributeSum}
\end{table*}

Image generation has been a widely-studied research problem in computer vision, with many competitive generative models proposed over the last decade, such as generative adversarial networks (GANs) \cite{NIPS2014_gan, arjovsky2017-wasserstein, karras-progressive-gan, brock2018-large, karras2019style} and variational autoencoders (VAE) \cite{kingma2013auto, pmlr-v32-rezende14, rezende2015variational, NEURIPS2019_vq-vae}.
Recently, diffusion models \cite{sohl2015deep,ho2020denoising,song2020denoising,song2020score}, with their ability to generate high-quality and high-resolution images in different domains, have soon attracted wide research attention.

One important research direction regarding image generative models is the ability to \emph{disentangle} different aspects of the generated images, such as semantic contents and styles, which is crucial for image editing and style transfer. A generative model with a good disentanglement capability should satisfy the following two desirable properties. First, it should permit separate modification of one aspect \emph{without} changing other aspects. As an example shown in Fig.~\ref{fig:motivation}, in text-to-image generation, when the text input changes from ``a photo of person'' to ``a photo of person with smile'', the generative model should have the ability to modify just the expression of the person (\emph{i.e.,} from the top image to middle image in Fig.~\ref{fig:motivation}) without changing the person's identity (the bottom image in Fig.~\ref{fig:motivation}). Second, the parameters learned from modifying one image should transfer well to other similar images. For example, the optimal parameters that can add smile to one person should also work for images of different people with different genders and races.

Previous studies have discovered that GANs are inherently endowed with a strong disentanglement capability. Specifically, it is found that there exist certain directions in the latent space separately controlling different attributes. 
Therefore, by identifying these directions, \emph{e.g.,} via principal component analysis \cite{harkonen2020ganspace}, GAN can achieve effective disentanglement \emph{without} any re-training or fine-tuning. On the other hand, such an inherent disentanglement capability has yet to be found in diffusion models. Hence come our research questions: Do diffusion models also possess a disentanglement capability with the aforementioned nice properties? If so, how can we uncover it?

In this paper, we seek to answer these research questions. Our finding is that for stable diffusion model \cite{rombach2022high}, one of the diffusion models that can generate images based on an input text description, disentangled image modifications can be achieved by partial modifications in the text embedding space. In particular, if we fix the standard Gaussian noises introduced in the denoising process, and partially change the input text embedding from a neutral description (\emph{e.g.,} ``a photo of person'') to one with style (\emph{e.g.,} ``a photo of person with smile''), the generated image will also shift towards the target style without changing the semantic content. Based on this finding, we further propose a simple, light-weight algorithm, where we optimize the mixing weights of the two text embeddings under two objectives, a perceptual loss for content preservation and a CLIP-based style matching loss. The entire process only involves optimizing over around 50 parameters and does not fine-tune the diffusion model.

Our experiments show that the inherent disentanglement capability in stable diffusion model can already disentangle a wide range of concepts and attributes, ranging from global styles such as painting styles to local styles like facial expressions, as shown in Table~\ref{tab:attributeSum}. As shown in Fig.~\ref{fig:intro}, by learning the optimal mixing weights of the two descriptions, stable diffusion models can generate convincing image pairs that only modify the target attribute, and the optimal weights can generalize well to different images. The experiment results also show that our proposed image editing algorithm, without fine-tuning the diffusion model, can match or outperform the more sophisticated diffusion-model-based image-editing baselines that require fine-tuning. The findings of this paper can shed some light on how diffusion models work and how they can be applied to image editing tasks.

\section{Related Works}

\noindent \textbf{Disentanglement in Generative Models:~}
The ability to disentangle different attributes is a key desired property of generative models.
Previous work that studies disentanglement mainly aims to learn parameters that allow modifications on a target aspect without changing other aspects, and the learned parameters should generalize to different images~\cite{karras2019style,achille2018emergence,eastwood2018framework}.
For pre-trained GANs~\cite{brock2018-large,karras2019style,karras2020analyzing,karras2021alias}, it has been shown that the disentanglement can be achieved by moving towards particular directions in its latent space~\cite{shen2020interfacegan,shen2020interpreting,harkonen2020ganspace,shen2021closed}, which will lead to attribute-only changes~\cite{abdal2019image2stylegan,abdal2020image2stylegan++,patashnik2021styleclip}.
Multiple methods have been proposed to discover these latent directions, which leverage auxiliary classifiers~\cite{shen2020interfacegan,shen2020interpreting}, principal component analysis~\cite{harkonen2020ganspace}
, contrastive learning~\cite{ren2021learning}, and information maximization~\cite{chen2016infogan}. Besides GANs, disentanglement has also been studied in VAE and flow-based models~\cite{kim2018disentangling,paige2017learning}. Recently, two works study disentanglement in diffusion models.
The first work disentangles attributes by learning a shift in the embedding space of an intermediate layer of U-Net \cite{kwon2022diffusion,ronneberger2015u}, such that applying the shift satisfies the disentanglement criteria.
Particularly, they use a neural network to generate such shifts.
However, disentangling in the hidden layer representation
of U-Net might be sub-optimal, as can be observed that their method struggles at disentangling holistic styles of the image.
Moreover, their search space and number of parameters are much larger than ours.
By contrast, we consider the text embedding space, which is more natural for text-to-image diffusion models and achieves comparable or better results with only \e{1.2\%} parameters of theirs.
Another work \cite{preechakul2022diffusion} trains an encoder to generate an image-specific representation, which is later used as input to diffusion models to reconstruct the original image. Disentanglement is done by finding corresponding directions in this representation space similar to methods in GANs \cite{shen2020interfacegan, shen2020interpreting}. However, their method requires re-training a diffusion model from scratch, whereas we fix the pre-trained diffusion model.

\vspace*{0.05in}
\noindent \textbf{Diffusion Models:~}
Diffusion models~\cite{sohl2015deep,ho2020denoising,song2020denoising,song2020score} are a family of generative models that have achieved state-of-the-art performance in image synthesis and have advanced research in super-resolution~\cite{ho2021cascaded,saharia2022-image}, inpainting~\cite{saharia2022palette,lugmayr2022repaint}, density estimation~\cite{Kingma2021VariationalDM}, video synthesis~\cite{ho2022-video,ho2022imagen}, and areas beyond computer vision~\cite{kong2020-diffwave, chen2020-wavegrad, austin2021-structured, tashiro2021-csdi, jing2022-torsional}.
Building on top of diffusion models, various methods have been proposed to control the generation process through external models~\cite{dhariwal2021diffusion, liu2021more} or additional inputs~\cite{ho2022-classifier}.
One type of conditional generation models is the text-to-image diffusion models~\cite{nichol2021-glide,ramesh2022hierarchical,saharia2022photorealistic,rombach2022high}, which take text descriptions as inputs and generate images that match the text descriptions.
Due to the expressiveness of text and superior generation quality of diffusion models, these models allow unprecedented control over generated images and have inspired many novel applications.

\vspace*{0.05in}
\noindent \textbf{Image Editing:~} Image editing is a widely-studied task \cite{zhu2017unpaired,hertzmann2001image}. Many GAN-based editing works~\cite{bau2020semantic,patashnik2021styleclip,jo2019sc,li2020manigan,liu2020describe} have demonstrated strong controllability. Recently, 
diffusion models have been broadly adapted to image editing task~\cite{hertz2022prompt,meng2021sdedit,liu2022compositional,couairon2022diffedit,li2022efficient}. 
With the CLIP encoder~\cite{radford2021learning} that bridges text and image, generation process can be guided by arbitrary text descriptions~\cite{liu2021more}.
To preserve the contents in a local region, \cite{avrahami2022blended} relies on an auxiliary mask, such that contents in the unmasked region are largely kept unchanged during generation.
Moreover, some works~\cite{kawar2022imagic,ruiz2022dreambooth,gal2022image} propose to invert the input image to find text embeddings that can synthesize the same object but in different scenes and views.
Although these works have demonstrated successful edits, there are two limitations. First, most of the methods require fine-tuning diffusion models~\cite{kim2022diffusionclip,kawar2022imagic,ruiz2022dreambooth}. For each editing task, they have to fine-tune and store the whole diffusion model, making them unscalable to a large amount of edits. Second, many methods rely on auxiliary inputs such as image masks~\cite{avrahami2022blended,ackermann2022high,avrahami2022blended2} or multiple examples of the edited object~\cite{ruiz2022dreambooth,gal2022image}, which are not always available.
Besides, only editing masked region may cause incoherence between masked and unmasked regions.
In this work, leveraging the disentanglement in stable diffusion, we propose to perform image editing without auxiliary inputs and the need to fine-tune diffusion models, which is more practical to use.

\section{Attribute Disentanglement in Stable Diffusion Models}
In this section, we will explore the disentanglement properties inherent in diffusion models, and then propose an approach for disentangled image modification and editing utilizing these properties.

\subsection{Preliminaries on Diffusion Models}
\label{sec:ddim}

We first provide a brief overview of the denoising diffusion implicit model (DDIM)~\cite{song2020denoising} conditioned on input text descriptions that is used in the stable diffusion model \cite{rombach2022high}, which we will be primarily studying in this paper.
Given a text embedding, denoted as \e{\bm c}, the goal of the text-conditioned DDIM is to generate an image, denoted as \e{\bm X_0}, that conforms to the text description. DDIM defines a set of noisy images, \e{\bm X_{1:T}}, by adding Gaussian noise to \e{\bm X_0} according to a predefined diffusion process. Each \e{\bm X_t} is corrupted with a larger noise than \e{\bm X_{t-1}}, and \e{\bm X_T} is very close to standard Gaussian noise.
The generation process of DDIM tries to denoise from \e{\bm X_T} all the way back to \e{\bm X_0}. Specifically, in the first step, \e{\bm X_T} is randomly drawn from a standard Gaussian distribution.
Then, each \e{\bm X_{t-1}} is inferred from \e{\bm X_{t}} via the following denoising process:
\begin{equation}
\small
    \bm X_{t-1} = \gamma_{t0} \bm X_t + \gamma_{t1} \bm \epsilon_{\theta}(\bm X_t, t, \bm c_t),
    \label{eq:denoise}
\end{equation}
where \e{\bm c_t} is the text embedding used at step \e{t}. In most common settings, generating one image only requires one text description \e{\bm c}, so \e{\bm c_t = \bm c} for all \e{t}. Here we make it dependent upon \e{t} to accommodate the discussions in the following subsections. \e{\bm \epsilon_{\theta^*}(\bm X_t, t, \bm c_t)} is a pre-trained denoising network that infers \e{\bm X_0} given the input of \e{\bm X_t} and \e{\bm c_t}. The parameters of the denoising network, \e{\theta^*}, are considered fixed throughout this section. \e{\gamma_{t0}} and \e{\gamma_{t1}} are defined as
\begin{equation}
\small
    \gamma_{t0} = \sqrt{\frac{\alpha_{t-1}}{\alpha_t}}, \quad \gamma_{t1} = \sqrt{1 - \alpha_{t-1}} - \sqrt{\frac{\alpha_{t-1}}{\alpha_t} - \alpha_{t-1}},
\end{equation}
and \e{\alpha_{0:T}} are hyperparameters that govern diffusion process.

Note that in the general DDIM framework, a Gaussian noise is added to each denoising step in Eq.~\eqref{eq:denoise}, but we follow the convention \cite{kim2022diffusionclip} to set the variance to 0 for better stability. Therefore, the generated image is a \emph{deterministic} function of initial random noise \e{\bm X_T} and the text descriptions \e{\bm c_{1:T}}. Thus we introduce the following notation,
\begin{equation}
    \small
    \bm X_0 = \bm g(\bm X_T, \bm c_{1:T}),
    \label{eq:summary}
\end{equation}
to summarize the image generation process. The stable diffusion model follows the same setup, except that the diffusion process is performed on a hidden embedding space, so Eq.~\eqref{eq:summary} can also summarize its generation process.

\subsection{The Disentanglement Properties}\label{sec3.2}
\label{subsec:observation}
\begin{figure}[t]
    \centering
    \includegraphics[width=0.9\columnwidth]{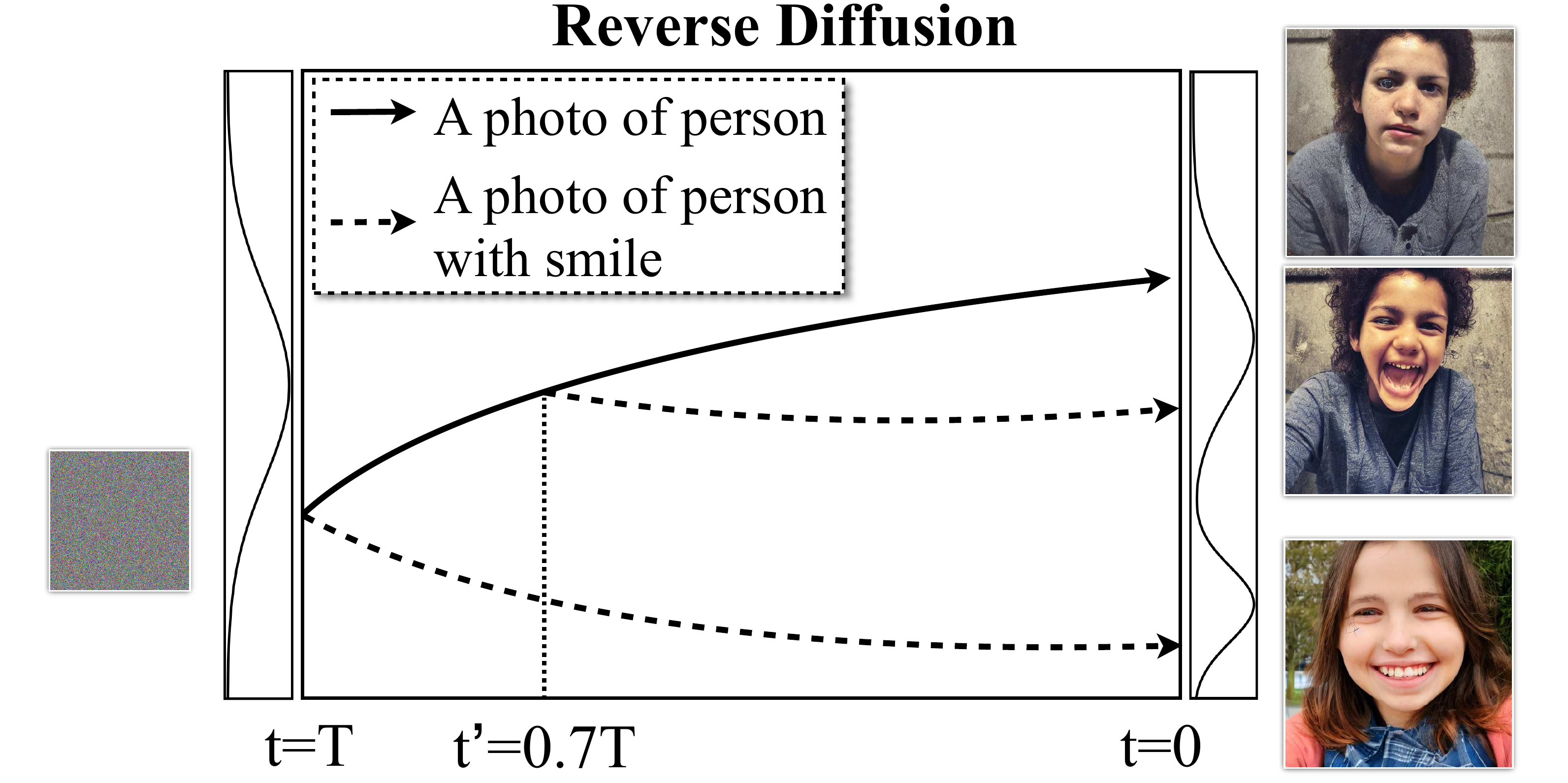}
    \vspace*{-0.075in}
    \caption{\textbf{The disentanglement property of stable diffusion models.} The top image is generated conditioned on  ``a photo of person''. The bottom image is generated with all descriptions replaced with ``a photo of person with smile'', and changes the person's identity. The middle image is generated by partially replacing descriptions at later steps,
    and maintains the person's identity.}
    \vspace*{-0.125in}
    \label{fig:motivation}
\end{figure}

Now we are ready to study whether the stable diffusion model is inherently capable of disentangling styles from semantic content. For concreteness, we will present our findings based on one specific example, but the findings are consistent across different cases.

Consider two text embeddings, \e{\bm c^{(0)}} and \e{\bm c^{(1)}}. \e{\bm c^{(0)}} is the embedding of a style-neutral description, ``a photo of person'', and \e{\bm c^{(1)}} is the embedding of a description with an explicit style, ``a photo of person with smile''. When \e{\bm c^{(0)}} is fed to the model, the generated image is a person with neutral expressions (top image in Fig.~\ref{fig:motivation}). We investigate whether the model can generate an image of the \emph{same person} with only the facial expression changed, when we fix \e{\bm X_T} (hence controlling all the randomness) and replace \e{\bm c^{(0)}} with \e{\bm c^{(1)}}.

\vspace*{0.05in}
\noindent \textbf{Case 1: Full Replacement.} 
In our first attempt, we replaced text embeddings at all denoising steps with \e{\bm c^{(1)}}. The resulting generated image undesirably changes the identity of the person, as shown in the bottom image in Fig.~\ref{fig:motivation}. 

\vspace*{0.05in}
\noindent \textbf{Case 2: Partial Replacement.}
In our second attempt, we replaced the text embeddings to \e{\bm c^{(1)}} only at the later denoising steps, \emph{i.e.,} 
\begin{equation}
    \small
    \bm c_t = \bm c^{(1)}, \forall t \leq t', \quad \bm c_t = \bm c^{(0)}, \forall t > t',
\end{equation}
where \e{t'=0.7T} in this example. In this case, the image can successfully maintain the identity of the person while only changing the smile, as shown in the middle image in Fig.~\ref{fig:motivation}.

As can be concluded, there exists an inherent disentanglement capability in stable diffusion models, which can be triggered by partially replacing the text embeddings, but not with full replacement. We have performed the same experiments on more objects and styles and the observations are consistent, as shown in Appendix~\ref{3.2disentangle}.

\subsection{Optimizing for Disentanglement}\label{optimization}
\label{sec:optimize}

\begin{figure*}
    \centering
    
    \includegraphics[width=0.9\textwidth]{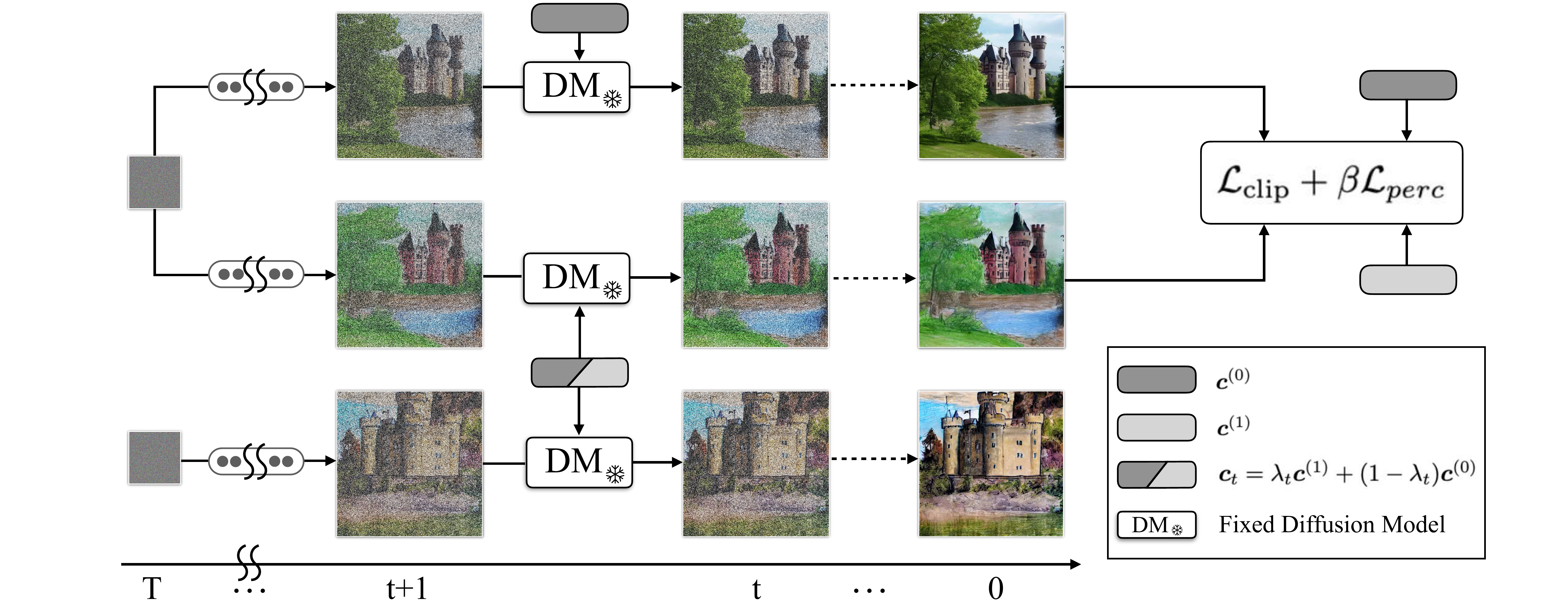}
    \vspace*{-0.05in}
    \caption{
    \textbf{Overview of our method that finds optimal text embedding for disentanglement.}
    In this example, \e{\bm c^{(0)}} is the embedding of ``A castle'', and \e{\bm c^{(1)}} is the embedding of ``A children drawing of castle''.
    \textbf{First two rows:} optimization process that finds the best soft combination of \e{\bm c^{(0)}} and \e{\bm c^{(1)}}, such that the modified image (the second row) changes the attribute without affecting other contents.
    \textbf{Last row:} the learned text embedding can be directly applied to a new image, which leads to the same editing effect.
    }
    \vspace*{-0.15in}

    \label{fig:method}
\end{figure*}

In Sec.~\ref{subsec:observation}, we have seen one way of realizing a disentangled generation of image using \e{\bm c^{(0)}} and \e{\bm c^{(1)}}. However, this is not necessarily the optimal way of combining the two embeddings in terms of disentanglement. In this subsection, we will propose a principled and tractable optimization scheme to combine a given pair of \e{\bm c^{(0)}} and \e{\bm c^{(1)}} to achieve the best disentanglement.

The key relaxation for the optimization framework is the \emph{soft combination} of the text embeddings. Specifically, instead of feeding either \e{\bm c^{(0)}} or \e{\bm c^{(1)}} at each denoising step \e{t}, we feed a soft combination of the two, namely
\begin{equation}
    \small
    \bm c_t = \lambda_t \bm c^{(1)} + (1-\lambda_t) \bm c^{(0)} \equiv \bm c^{(\lambda)}_t,
\end{equation}
where \e{\lambda_t} is a learnable combination weight. The soft combination offers a much richer representation power, with both cases discussed in Sec.~\ref{subsec:observation} being its special cases (by setting \e{\lambda_t} to either 1 or 0).

Given a random noise \e{\bm X_T} and the text description pair \e{\bm c^{(0)}} and \e{\bm c^{(1)}}, the optimization procedure for \e{\lambda_{1:T}} is as follows. First, two images are generated, a style-neutral one generated with \e{\bm c^{(0)}} and the other generated with \e{\bm c^{(\lambda)}_{1:T}}:
\begin{equation}
    \small
    \bm X_0^{(0)} = \bm g(\bm X_T, \bm c^{(0)}), \quad \bm X_0^{(\lambda)} = \bm g(\bm X_T, \bm c^{(\lambda)}_{1:T}).
    \label{eq:generation}
\end{equation}
Our goal is to find an optimal \e{\lambda_{1:T}} such that \e{\bm X_0^{(\lambda)}} maintains the same semantic content as \e{\bm X_0^{(0)}} but conforms to the style described in \e{\bm c^{(1)}}, which is achieved by solving the following optimization problem similar to \cite{patashnik2021styleclip,kwon2022diffusion}:
\begin{equation}
\small
    \min_{\lambda_{1:T}} \mathcal{L}_{\textrm{clip}}(\bm X_0^{(0)}, \bm X_0^{(\lambda)}, \bm c^{(0)}, \bm c^{(1)}) + \beta \mathcal{L}_{perc}(\bm X_0^{(0)}, \bm X_0^{(\lambda)}).
\label{eq:loss}
\end{equation}
\e{\beta} is the hyperparameter that balances the two loss terms. \e{\mathcal{L}_{\textrm{clip}}} is the directional CLIP loss~\cite{gal2022stylegan} that encourages the image change from \e{\bm X_0^{(0)}} to \e{\bm X_0^{(\lambda)}} matches the text change from \e{\bm c^{(0)}} to \e{\bm c^{(1)}} in the CLIP embedding space.\footnote{In practice, we use a different text embedder for $\mathcal{L}_{clip}$, but for brevity we still use $\bm c^{(0)}$ and $\bm c^{(1)}$ to denote the embeddings.} \e{\mathcal{L}_{perc}} is the perceptual loss \cite{johnson2016perceptual} that prevents drastic changes in semantic content:
\begin{equation}
    \small
    \mathcal{L}_{perc}(\bm X_0^{(0)}, \bm X_0^{(\lambda)}) = \Vert \bm h(\bm X_0^{(0)}) - \bm h(\bm X_0^{(\lambda)}) \Vert_1,
\end{equation}
where \e{\bm h(\cdot)} denotes a perceptual network that encodes a given image. In stable diffusion models, the number of denoising steps can be as few as 50 and so are the corresponding \e{\lambda_{1:T}}, so this optimization problem involves very few parameters and does not need to fine-tune the diffusion model itself.

\subsection{Extension to Image Editing}
\label{sec:edit}
With the disentangled image modification algorithm developed, we can now extend the approach to achieve disentangled image editing. 
The only difference of the image editing setting compared to the settings in the previous subsections is that rather than having the diffusion model generate the neutral image conditioned on \e{\bm c^{(0)}}, the neutral image is now externally given, which we denote as \e{\bm I} to show the distinction. Therefore, if we could find a value for initial random variable \e{\bm X_T} such that the stable diffusion model can generate exactly the same image as \e{\bm I} when conditioned on \e{\bm c^{(0)}}, we can then use the same approach in Sec.~\ref{sec:optimize} for the image editing task.

To this end, we adapt the image inversion approach proposed in \cite{kim2022diffusionclip, hertz2022prompt}, which recursively generates a set of noisy images, \e{\hat{\bm X}_{1:T}} based on \e{\bm I} as follows:
\begin{equation}
    \small
    \hat{\bm X}_0 = \bm I, \quad \hat{\bm X}_{t+1} = \gamma'_{t0} \hat{\bm X}_t + \gamma'_{t1} \bm \epsilon_{\bm \theta} (\hat{\bm X}_t, t, \bm c^{(0)}),
\end{equation}
where
\begin{equation}
\small
    \gamma'_{t0} = \sqrt{\frac{\alpha_{t+1}}{\alpha_t}}, \quad \gamma'_{t1} = \sqrt{1 - \alpha_{t+1}} - \sqrt{\frac{\alpha_{t+1}}{\alpha_t} - \alpha_{t+1}}.
\end{equation}
It has been shown \cite{kim2022diffusionclip} that by setting \e{\bm X_T = \hat{\bm X}_T} and following through the generation process in Eq.~\eqref{eq:denoise} with \e{\bm c_t = \bm c^{(0)}}, the resulting \e{\bm X_{0:T-1}} would satisfy \e{\bm X_t \approx \hat{\bm X}_t, \forall t \in [0, T-1]}, and therefore \e{\bm X_0 \approx \bm I}.

To further close the approximation gap, we introduce a new diffusion process, where the approximation error is added as a correction term. Formally, the new diffusion process starts with \e{\tilde{\bm X}_T = \hat{\bm X}_T}, and then
\begin{equation}
    \small
    \tilde{\bm X}_{t-1} = \gamma_{t0} \tilde{\bm X}_t + \gamma_{t1} \bm \epsilon_{\theta}(\tilde{\bm X_t}, t, \bm c_t) + \bm E_t,
\end{equation}
where the \e{\bm E_t} is the correction term defined as
\begin{equation}
    \small
    \bm E_t = \hat{\bm X}_{t-1} - \gamma_{t0} \hat{\bm X}_t - \gamma_{t1} \bm \epsilon_{\theta}(\hat{\bm X_t}, t, \bm c^{(0)}).
\end{equation}
Again, for notational brevity, we summarize this new generation process as
\begin{equation}
\small
    \tilde{\bm X}_0 = \tilde{\bm g}(\tilde{\bm X}_T, \bm c_{1:T}, \bm E_{1:T}).
\end{equation}
It can be easily shown that \e{\tilde{\bm g}(\hat{\bm X}_T, \bm c^{(0)}, \bm E_{1:T}) = \bm I}.

Now that we have developed a generation process that can reconstruct \e{\bm I}, we can now follow the same procedure in Sec.~\ref{sec:optimize} to perform image editing, with the image generation in Eq.~\eqref{eq:generation} replaced with
\begin{equation}
    \small
    \tilde{\bm X}^{(0)}_0 = \tilde{\bm g}(\hat{\bm X}_T, \bm c^{(0)}, \bm E_{1:T}) = \bm I, ~\tilde{\bm X}^{(\lambda)}_0 = \tilde{\bm g}(\hat{\bm X}_T, \bm c_{1:T}^{(\lambda)}, \bm E_{1:T}).
\end{equation}
It is worth emphasizing that when generating \e{\tilde{\bm X}^{(\lambda)}_0}, the error correction terms \e{\bm E_{1:T}} are still fixed to the ones computed for reconstructing \e{\bm I}. To further enhance the quality of the edited image, we adopt the re-diffusion approach in \cite{meng2021sdedit}.

\begin{figure*}
\centering
\includegraphics[width=\textwidth]{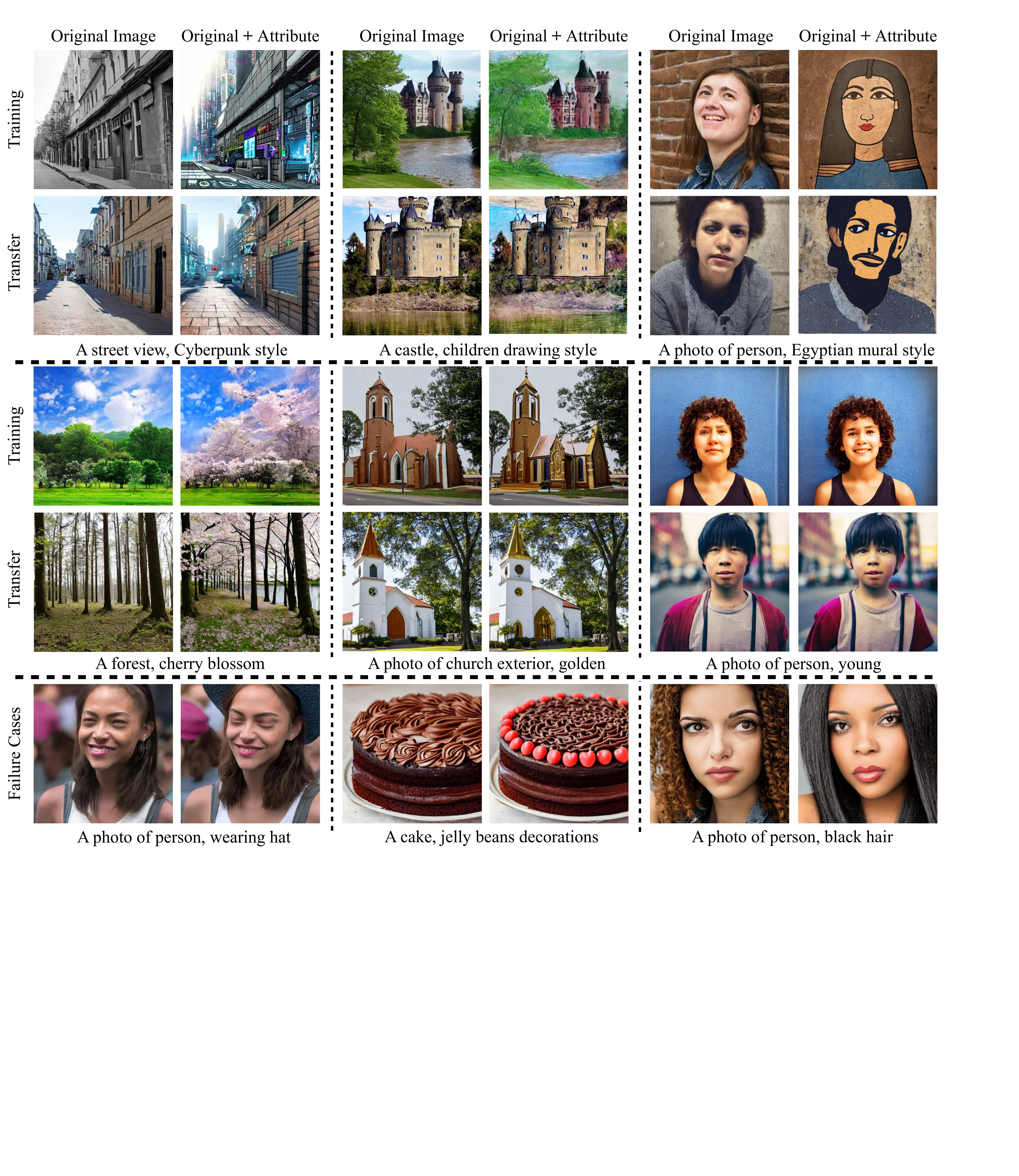}
\vspace{-2.1em}
\caption{
{\small \textbf{Example disentanglement results.}
Text description with style (\e{\bm c^{(1)}}) is shown below each row, which consists of the style-neutral text description (\e{\bm c^{(0)}}) and target attribute description, separated by comma.
\textbf{Top two panels:} successful cases.
\textbf{Bottom panel:} failure cases.
Within each attribute, \textbf{first row:} results on optimization images; \textbf{second row:} results of transferring to unseen images; \textbf{left column:} source images; \textbf{right column:} modified images.
More examples can be found in Appendix \ref{4.1more_example}.}}
\vspace*{-0.1in}
\label{fig:exp1}
\end{figure*}

\section{Experiments}\label{exp}

We will perform experiments to explore the inherent disentanglement capability in the stable diffusion model and evaluate the performance of our method.

\vspace*{0.05in}
\noindent \textbf{Implementation Details:~}
For all experiments, we use the diffusion model \texttt{stable-diffusion-v1-4}~\cite{rombach2022high}, which is pre-trained on \texttt{laion} dataset~\cite{schuhmann2021laion}.
The pre-trained model is \textit{frozen} throughout all experiments, and we keep the default hyperparameters of the model.
All images generated by our method are in size $512\times 512$. 
We use a variant of the DDIM sampler~\cite{liu2022pseudo} to synthesize images with 50 total backward diffusion steps.
To optimize \e{\lambda_{1:T}}, we use Adam~\cite{kingma2014adam} optimizer with learning rate 0.03. To balance two loss terms,
we set \e{\beta} to 0.05 for all human face experiments and 0.03 for all experiments on scenes and buildings. Finally, when editing real images, the number of re-diffusion steps is 20.
More information on the hyperparameters and optimization is in Appendix \ref{hyperparameters}.

\subsection{Exploring the Disentanglement Capability}
\label{exp1}

Since the foundation of our method hinges on the inherent disentanglement capability in the stable diffusion model,
we would like to first explore the strength and the limit of this capability. In particular, the research question we would like to answer is what are the attributes and objects that the stable diffusion model can inherently disentangle well, and what cannot.  To this end, we perform a comprehensive qualitative study where we first compile a list of objects and attributes that have been tested upon in existing work on image disentanglement \cite{kwon2022diffusion,shen2020interfacegan,harkonen2020ganspace}, as shown in Table~\ref{tab:attributeSum}. For each object-attribute pair in the list, we craft a text description pair, the neutral description \e{\bm c^{(0)}} and the description with style \e{\bm c^{(1)}}. In particular, \e{\bm c^{(0)}} includes just the name of the object, \emph{e.g.,} ``A castle'', and \e{\bm c^{(1)}} is constructed by appending the description of the attribute to \e{\bm c^{(0)}}, separated by comma (\emph{e.g.,} ``A castle, children drawing style''). Next, we generate five style-neutral images for each \e{\bm c^{(0)}} and
use the optimization method in Sec.~\ref{sec:optimize} to disentangle the attribute from these five images (more details in Appendix~\ref{text-list}).

\begin{figure*}[t!]
    \centering
    \includegraphics[width=0.95\textwidth]{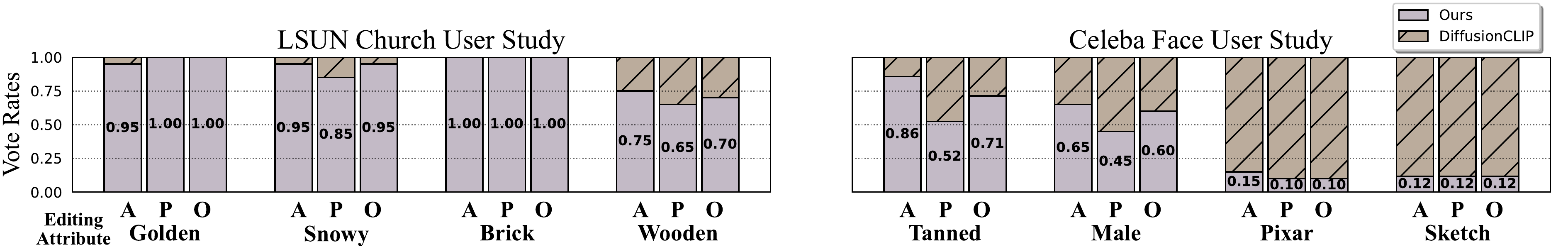}
    \vspace*{-0.1in}
    \caption{\textbf{Human evlautions} on \textbf{A (attribute similarity)}, \textbf{P (content preservation)}, and \textbf{O (overall quality)}. The number shown in each bar indicates the percentage of users who vote for our method is better than the baseline for the corresponding aspect.}    
    \vspace*{-0.1in}
    \label{fig:human-study}
\end{figure*}

\vspace*{0.05in}
\noindent \textbf{Results:}
The results are shown in Table~\ref{tab:attributeSum}, and some generated image pairs with their text descriptions are shown in Fig.~\ref{fig:exp1}, where the top two panels show some successful cases. In each panel, the first row (marked ``Training'') shows the results where the optimal \e{\lambda_{1:T}} are learned specific to that image pair; the second row (marked ``Transfer'') shows the results of applying the learned \e{\lambda_{1:T}} to a new image with the same object and target attribute. As can be observed, using our method, the stable diffusion model, without any re-training, can disentangle a surprisingly wide range of objects and attributes already. In particular, our method is strong at disentangling global styles that cover largely the whole image, such as scenery styles, drawing styles, and architecture materials, where the scenery layout or building structure is largely maintained while only the target attribute is modified. Our method can also disentangle many local attributes like facial expressions. Besides, the learned \e{\lambda_{1:T}} has a great transferability to unseen images, thus satisfying both disentanglement criteria in Sec.~\ref{sec:intro}.

On the other hand, stable diffusion has difficulties disentangling attributes that involve small objects, \emph{e.g.,} adding small accessories, as shown in the bottom panel of Fig.~\ref{fig:exp1}. When the target attribute is added, the model tends to also change some other correlated attributes, such as the style of the cake or the identity of the person, which may be ascribed to the model's weaker control of finer-grained details. Nevertheless, these results suggest that the
disentanglement capability in an unmodified stable diffusion model may be stronger than previously revealed, which provides a good justification for our method that applies it as is to image editing. More examples are shown in Appendix \ref{4.1more_example}.

\subsection{Evaluation on Disentangled Image Editing}\label{human-study}

In this section, we will evaluate our proposed method on the image editing task as described in Sec.~\ref{sec:edit}.

\vspace*{0.05in}
\noindent \textbf{Configurations:~} Since there is no ground-truth for the image editing task, we conduct a subjective evaluation on Amazon Mechanical Turk. Specifically, we use the Celeb-A \cite{liu2015faceattributes} and LSUN-church \cite{yu2015lsun}. For each dataset, we use the first 20 images as the source image and perform 4 types of the edit used in~\cite{kim2022diffusionclip} (\emph{i.e.,} tanned, male, sketch, pixar for human faces; and golden, wooden, snowy, red brick for churches). The main baseline is \textsc{DiffusionCLIP}, which is a state-of-the-art diffusion-based image editing approach that requires fine-tuning. Each subject was presented with a pair of images edited by the two methods from the same source image in a randomized order and asked three questions regarding the editing quality: (1) (\emph{attribute similarity}) which image better incorporates the target attribute in a natural way; (2) (\emph{content preservation}) which image better preserves other contents; and (3) (\emph{overall}) which image has better overall quality. More details are in Appendix~\ref{human_study}.

\vspace*{0.05in}
\noindent \textbf{Subjective Evaluation Results:~} 
Fig.~\ref{fig:human-study} shows the results of the percentage of answers that chose our method in each of the three questions and in each target attribute. As can be observed, in 6 out of the 8 attributes, our method outperforms the baseline in almost all three aspects, demonstrating the high quality of its disentanglement. We discovered a common failure case in \textsc{DiffusionCLIP} where the attributes are so over-optimized that some artifacts are introduced and some irrelevant parts of the image are modified (examples in Appendix~\ref{human_study}). Our method, with its light-weight optimization, can avoid the over-optimization problem.  On the other hand, our method is less competitive in human-related editing, underperforming the baseline in two attributes, ``sketch'' and ``pixar''. One potential cause is that the diffusion model in \textsc{DiffusionCLIP} is specifically trained to generate human images. Nevertheless, these results show that the inherent disentanglement capability in stable diffusion can enable powerful image editing that can even outperform the baseline that requires fine-tuning.

\vspace*{0.05in}
\noindent \textbf{Qualitative Comparison with More Baselines:~} 
\begin{figure}[t]
    \centering
    \includegraphics[width=0.99\columnwidth]{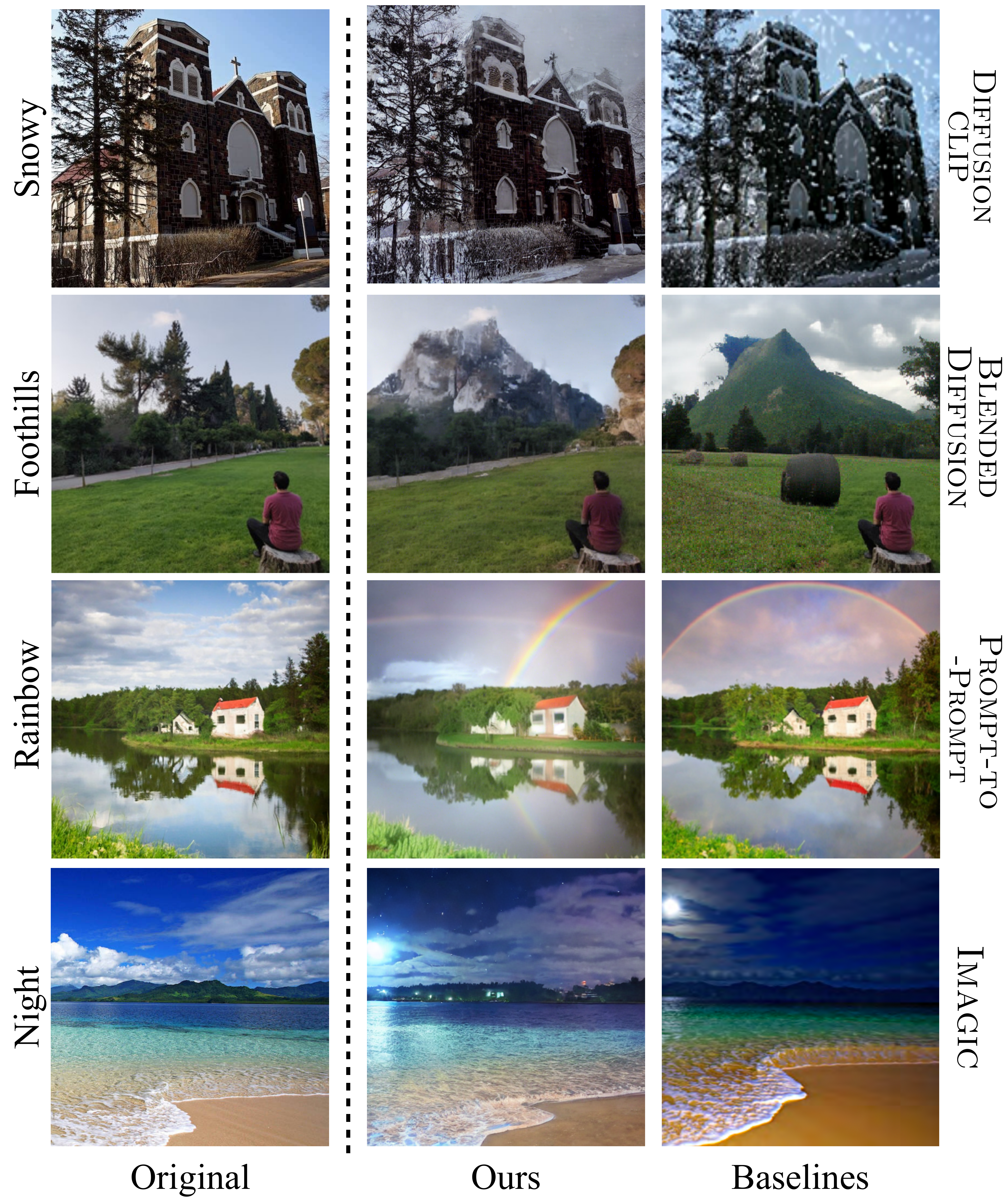}
    \caption{\textbf{Visual comparisons.} The source images and the corresponding baseline results are taken from the original papers.  Each row indicates the comparison to one particular baseline. }
    \vspace*{-0.1in}
    \label{fig:qualitative}
\end{figure}
Besides \textsc{DiffusionCLIP}, we have identified three other baselines that use diffusion models for image editing, which are \textsc{Blended-Diffusion}~\cite{avrahami2022blended}, \textsc{Prompt-to-Prompt}~\cite{hertz2022prompt}, and \textsc{Imagic}~\cite{kawar2022imagic}. However, these methods either have not released code by the time of our submission or require auxiliary labels that are unavailable, so we could not include them in our subjective evaluation. Nevertheless, we managed to generate some qualitative comparisons by collecting the source and edited images shown on their papers and using our method to perform the edit. The edited images are displayed side-by-side in Fig.~\ref{fig:qualitative}, with each row corresponding to one baseline.  As can be observed, our method maintains the semantic content better than \textsc{Blended Diffusion}, which undesirably changes the details such as the grass.  Our method performs relatively on par with \textsc{DiffusionCLIP} and \textsc{Prompt-to-Prompt} in terms of both attribute matching and content preservation.
Finally, compared with \textsc{Imagic}, our method arguably changes slightly more semantic content (\emph{e.g.,} adding stars) but the overall quality is very competitive.  To sum up, these results, and additional results in Appendix~\ref{compare_baseline_qualitative}, consistently verify the high quality of our edited images against each baseline's most representative results reported in their papers.

\subsection{Ablation Study}\label{ablation-main}

As described in Sec.~\ref{sec:optimize}, the optimal \e{\lambda_{1:T}} depends on specific values of \e{\bm c^{(0)}}, \e{\bm c^{(1)}}, and \e{\bm X^T}. We thus like to investigate whether our image editing algorithm is robust against variations in these inputs. To check the robustness against \e{\bm c^{(0)}} and \e{\bm c^{(1)}} in the image editing task, we generate different edited images from the same source image by varying the text descriptions. We study three types of variations.

\begin{figure}[t]
\centering
\includegraphics[width=0.99\columnwidth]{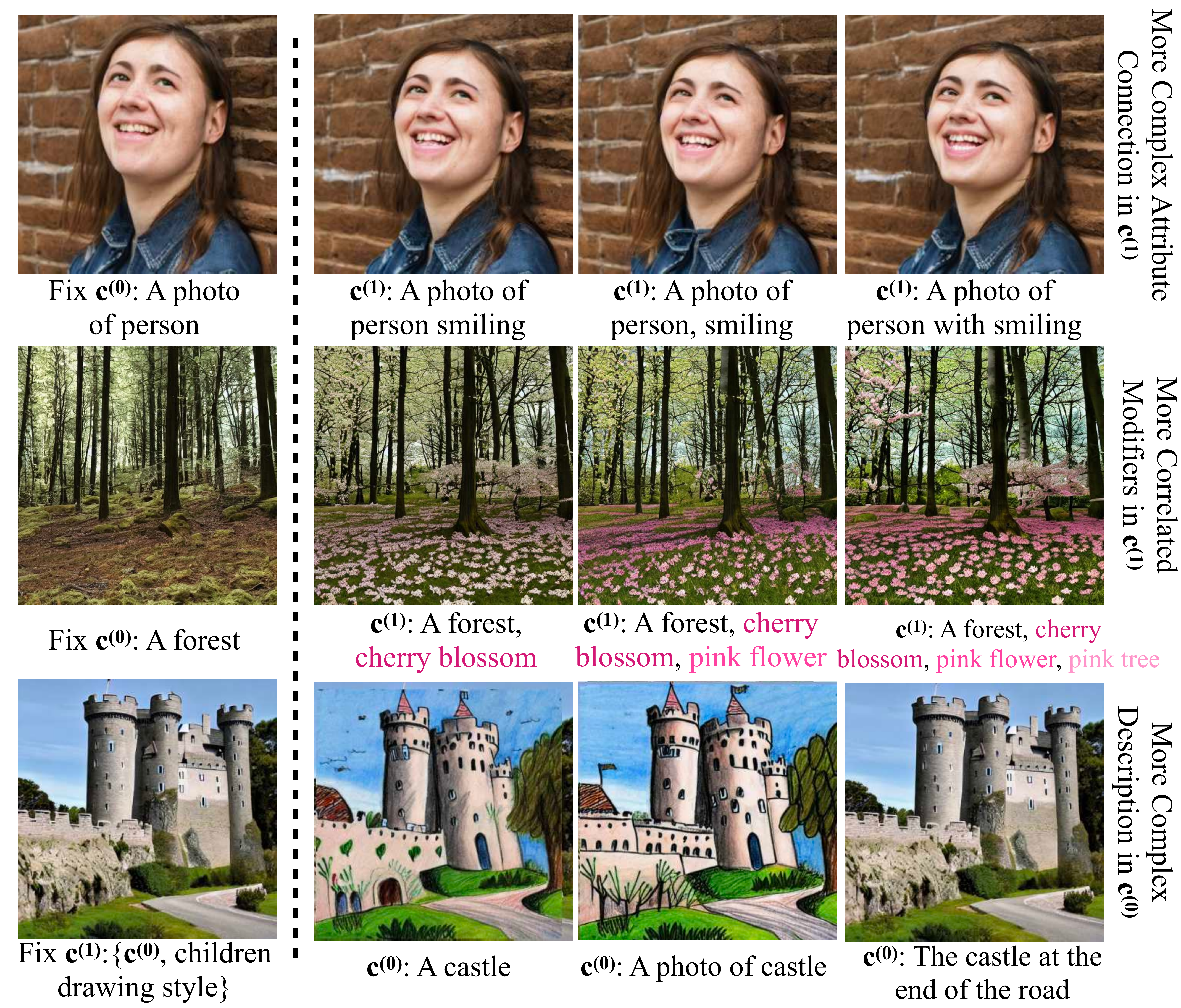}
\caption{\textbf{Text descriptions for image editing.} The original image is shown in the first column. On the right columns, we demonstrate the results when changing text descriptions \e{\bm c^{(0)}} and \e{\bm c^{(1)}}.}
\vspace*{-0.15in}
\label{fig:ablation-1}
\end{figure}

First, we study how the way that \e{\bm c^{(1)}} appends the target attribute description can affect the results. Specifically, we fix \e{\bm c^{(0)}} and the target attribute, and generate three  \e{\bm c^{(1)}}'s with different ways to concatenate the target attribute: direct concatenation, concatenation separated by a comma, and concatenation separated by ``with'', as illustrated in the top row of Fig.~\ref{fig:ablation-1}. As can be seen, the resulting edit images are hardly impacted by such variations.

Second, we study whether varying the complexity of the target attribute description in \e{\bm c^{(1)}} would affect the results. To this end, we again fix \e{\bm c^{(0)}} and gradually add more modifiers describing the same target attribute to \e{\bm c^{(1)}}, as shown in the second row of Fig.~\ref{fig:ablation-1}. As can be seen, while having more modifiers can amplify the editing effects, having one modifier is sufficient to generate an effective edit.

Finally, we study whether variations in \e{\bm c^{(0)}} would affect the results. Specifically, we fix the target attribute description and generate three versions of \e{\bm c^{(0)}} that describe the same source object, a short description, a longer description by adding non-informative words, and a description generated by an image captioning model~\cite{wang2022unifying}.
As shown in the bottom row of Fig.~\ref{fig:ablation-1}, the image editing is largely consistent in almost all cases, but fails with the \e{\bm c^{(0)}} generated by the image captioning model. One possible cause is that the generated caption is usually lengthy and contains more details, which may overwhelm the target attribute description.

More image examples are shown in Appendix~\ref{text-influence}, and the overall conclusion is that the edited results are highly robust against most variations in the text descriptions. We also perform some experiments that show strong robustness against variations in the images on which \e{\lambda_{1:T}} are learned, which is presented in Appendix~\ref{optimization-image}.

\section{Conclusion}
In this work, we study the disentanglement property in the stable diffusion model.
We find that stable diffusion inherently has the disentanglement capability, and it can be triggered by partially replacing the text embedding from a style-neutral description to one with desired style. 
Motivated by this finding, we propose a simple and light-weight disentanglement algorithm where the combination weights of the two text embeddings are optimized for style matching and content preservation. With only 50 parameters being optimized, our method demonstrates generalizable disentanglement ability and outperforms sophisticated baselines that require fine-tuning on image editing task.

{\small
\bibliographystyle{ieee_fullname}
\bibliography{egbib}

\begin{thebibliography}{10}\itemsep=-1pt

\bibitem{abdal2019image2stylegan}
Rameen Abdal, Yipeng Qin, and Peter Wonka.
\newblock Image2stylegan: How to embed images into the stylegan latent space?
\newblock In {\em ICCV}, 2019.

\bibitem{abdal2020image2stylegan++}
Rameen Abdal, Yipeng Qin, and Peter Wonka.
\newblock Image2stylegan++: How to edit the embedded images?
\newblock In {\em CVPR}, 2020.

\bibitem{achille2018emergence}
Alessandro Achille and Stefano Soatto.
\newblock Emergence of invariance and disentanglement in deep representations.
\newblock {\em The Journal of Machine Learning Research}, 2018.

\bibitem{ackermann2022high}
Johannes Ackermann and Minjun Li.
\newblock High-resolution image editing via multi-stage blended diffusion.
\newblock {\em arXiv preprint arXiv:2210.12965}, 2022.

\bibitem{arjovsky2017-wasserstein}
Martin Arjovsky, Soumith Chintala, and Léon Bottou.
\newblock Wasserstein gan, 2017.

\bibitem{austin2021-structured}
Jacob Austin, Daniel~D. Johnson, Jonathan Ho, Daniel Tarlow, and Rianne van~den
  Berg.
\newblock Structured denoising diffusion models in discrete state-spaces, 2021.

\bibitem{avrahami2022blended2}
Omri Avrahami, Ohad Fried, and Dani Lischinski.
\newblock Blended latent diffusion.
\newblock {\em arXiv preprint arXiv:2206.02779}, 2022.

\bibitem{avrahami2022blended}
Omri Avrahami, Dani Lischinski, and Ohad Fried.
\newblock Blended diffusion for text-driven editing of natural images.
\newblock In {\em CVPR}, 2022.

\bibitem{bau2020semantic}
David Bau, Hendrik Strobelt, William Peebles, Jonas Wulff, Bolei Zhou, Jun-Yan
  Zhu, and Antonio Torralba.
\newblock Semantic photo manipulation with a generative image prior.
\newblock {\em arXiv preprint arXiv:2005.07727}, 2020.

\bibitem{brock2018-large}
Andrew Brock, Jeff Donahue, and Karen Simonyan.
\newblock Large scale gan training for high fidelity natural image synthesis,
  2018.

\bibitem{chen2020-wavegrad}
Nanxin Chen, Yu Zhang, Heiga Zen, Ron~J. Weiss, Mohammad Norouzi, and William
  Chan.
\newblock Wavegrad: Estimating gradients for waveform generation, 2020.

\bibitem{chen2016infogan}
Xi Chen, Yan Duan, Rein Houthooft, John Schulman, Ilya Sutskever, and Pieter
  Abbeel.
\newblock Infogan: Interpretable representation learning by information
  maximizing generative adversarial nets.
\newblock In {\em Neurips}, 2016.

\bibitem{couairon2022diffedit}
Guillaume Couairon, Jakob Verbeek, Holger Schwenk, and Matthieu Cord.
\newblock Diffedit: Diffusion-based semantic image editing with mask guidance.
\newblock {\em arXiv preprint arXiv:2210.11427}, 2022.

\bibitem{dhariwal2021diffusion}
Prafulla Dhariwal and Alexander Nichol.
\newblock Diffusion models beat gans on image synthesis.
\newblock In {\em Neurips}, 2021.

\bibitem{eastwood2018framework}
Cian Eastwood and Christopher~KI Williams.
\newblock A framework for the quantitative evaluation of disentangled
  representations.
\newblock In {\em ICLR}, 2018.

\bibitem{gal2022image}
Rinon Gal, Yuval Alaluf, Yuval Atzmon, Or Patashnik, Amit~H Bermano, Gal
  Chechik, and Daniel Cohen-Or.
\newblock An image is worth one word: Personalizing text-to-image generation
  using textual inversion.
\newblock {\em arXiv preprint arXiv:2208.01618}, 2022.

\bibitem{gal2022stylegan}
Rinon Gal, Or Patashnik, Haggai Maron, Amit~H Bermano, Gal Chechik, and Daniel
  Cohen-Or.
\newblock Stylegan-nada: Clip-guided domain adaptation of image generators.
\newblock {\em ACM Transactions on Graphics (TOG)}, 2022.

\bibitem{NIPS2014_gan}
Ian Goodfellow, Jean Pouget-Abadie, Mehdi Mirza, Bing Xu, David Warde-Farley,
  Sherjil Ozair, Aaron Courville, and Yoshua Bengio.
\newblock Generative adversarial nets.
\newblock In Z. Ghahramani, M. Welling, C. Cortes, N. Lawrence, and K.Q.
  Weinberger, editors, {\em Neurips}, 2014.

\bibitem{harkonen2020ganspace}
Erik H{\"a}rk{\"o}nen, Aaron Hertzmann, Jaakko Lehtinen, and Sylvain Paris.
\newblock Ganspace: Discovering interpretable gan controls.
\newblock In {\em Neurips}, 2020.

\bibitem{hertz2022prompt}
Amir Hertz, Ron Mokady, Jay Tenenbaum, Kfir Aberman, Yael Pritch, and Daniel
  Cohen-Or.
\newblock Prompt-to-prompt image editing with cross attention control.
\newblock {\em arXiv preprint arXiv:2208.01626}, 2022.

\bibitem{hertzmann2001image}
Aaron Hertzmann, Charles~E Jacobs, Nuria Oliver, Brian Curless, and David~H
  Salesin.
\newblock Image analogies.
\newblock In {\em Proceedings of the 28th annual conference on Computer
  graphics and interactive techniques}, 2001.

\bibitem{ho2022imagen}
Jonathan Ho, William Chan, Chitwan Saharia, Jay Whang, Ruiqi Gao, Alexey
  Gritsenko, Diederik~P Kingma, Ben Poole, Mohammad Norouzi, David~J Fleet,
  et~al.
\newblock Imagen video: High definition video generation with diffusion models.
\newblock {\em arXiv preprint arXiv:2210.02303}, 2022.

\bibitem{ho2020denoising}
Jonathan Ho, Ajay Jain, and Pieter Abbeel.
\newblock Denoising diffusion probabilistic models.
\newblock {\em Neurips}, 2020.

\bibitem{ho2021cascaded}
Jonathan Ho, Chitwan Saharia, William Chan, David~J Fleet, Mohammad Norouzi,
  and Tim Salimans.
\newblock Cascaded diffusion models for high fidelity image generation.
\newblock {\em arXiv preprint arXiv:2106.15282}, 2021.

\bibitem{ho2022-classifier}
Jonathan Ho and Tim Salimans.
\newblock Classifier-free diffusion guidance, 2022.

\bibitem{ho2022-video}
Jonathan Ho, Tim Salimans, Alexey Gritsenko, William Chan, Mohammad Norouzi,
  and David~J. Fleet.
\newblock Video diffusion models, 2022.

\bibitem{jing2022-torsional}
Bowen Jing, Gabriele Corso, Jeffrey Chang, Regina Barzilay, and Tommi Jaakkola.
\newblock Torsional diffusion for molecular conformer generation, 2022.

\bibitem{jo2019sc}
Youngjoo Jo and Jongyoul Park.
\newblock Sc-fegan: Face editing generative adversarial network with user's
  sketch and color.
\newblock In {\em ICCV}, 2019.

\bibitem{johnson2016perceptual}
Justin Johnson, Alexandre Alahi, and Li Fei-Fei.
\newblock Perceptual losses for real-time style transfer and super-resolution.
\newblock In {\em ECCV}, 2016.

\bibitem{karras-progressive-gan}
Tero Karras, Timo Aila, Samuli Laine, and Jaakko Lehtinen.
\newblock Progressive growing of gans for improved quality, stability, and
  variation, 2017.

\bibitem{karras2021alias}
Tero Karras, Miika Aittala, Samuli Laine, Erik H{\"a}rk{\"o}nen, Janne
  Hellsten, Jaakko Lehtinen, and Timo Aila.
\newblock Alias-free generative adversarial networks.
\newblock In {\em Neurips}, 2021.

\bibitem{karras2019style}
Tero Karras, Samuli Laine, and Timo Aila.
\newblock A style-based generator architecture for generative adversarial
  networks.
\newblock In {\em CVPR}, 2019.

\bibitem{karras2020analyzing}
Tero Karras, Samuli Laine, Miika Aittala, Janne Hellsten, Jaakko Lehtinen, and
  Timo Aila.
\newblock Analyzing and improving the image quality of stylegan.
\newblock In {\em CVPR}, 2020.

\bibitem{kawar2022imagic}
Bahjat Kawar, Shiran Zada, Oran Lang, Omer Tov, Huiwen Chang, Tali Dekel, Inbar
  Mosseri, and Michal Irani.
\newblock Imagic: Text-based real image editing with diffusion models.
\newblock {\em arXiv preprint arXiv:2210.09276}, 2022.

\bibitem{kim2022diffusionclip}
Gwanghyun Kim, Taesung Kwon, and Jong~Chul Ye.
\newblock Diffusionclip: Text-guided diffusion models for robust image
  manipulation.
\newblock In {\em CVPR}, 2022.

\bibitem{kim2018disentangling}
Hyunjik Kim and Andriy Mnih.
\newblock Disentangling by factorising.
\newblock In {\em ICML}, 2018.

\bibitem{kingma2014adam}
Diederik~P Kingma and Jimmy Ba.
\newblock Adam: A method for stochastic optimization.
\newblock In {\em ICLR}, 2015.

\bibitem{Kingma2021VariationalDM}
Diederik~P. Kingma, Tim Salimans, Ben Poole, and Jonathan Ho.
\newblock Variational diffusion models.
\newblock {\em ArXiv}, abs/2107.00630, 2021.

\bibitem{kingma2013auto}
Diederik~P Kingma and Max Welling.
\newblock Auto-encoding variational bayes.
\newblock {\em arXiv preprint arXiv:1312.6114}, 2013.

\bibitem{kong2020-diffwave}
Zhifeng Kong, Wei Ping, Jiaji Huang, Kexin Zhao, and Bryan Catanzaro.
\newblock Diffwave: A versatile diffusion model for audio synthesis, 2020.

\bibitem{kwon2022diffusion}
Mingi Kwon, Jaeseok Jeong, and Youngjung Uh.
\newblock Diffusion models already have a semantic latent space.
\newblock {\em arXiv preprint arXiv:2210.10960}, 2022.

\bibitem{li2020manigan}
Bowen Li, Xiaojuan Qi, Thomas Lukasiewicz, and Philip~HS Torr.
\newblock Manigan: Text-guided image manipulation.
\newblock In {\em CVPR}, 2020.

\bibitem{li2022efficient}
Muyang Li, Ji Lin, Chenlin Meng, Stefano Ermon, Song Han, and Jun-Yan Zhu.
\newblock Efficient spatially sparse inference for conditional gans and
  diffusion models.
\newblock {\em arXiv preprint arXiv:2211.02048}, 2022.

\bibitem{liu2022pseudo}
Luping Liu, Yi Ren, Zhijie Lin, and Zhou Zhao.
\newblock Pseudo numerical methods for diffusion models on manifolds.
\newblock In {\em ICLR}, 2022.

\bibitem{liu2022compositional}
Nan Liu, Shuang Li, Yilun Du, Antonio Torralba, and Joshua~B Tenenbaum.
\newblock Compositional visual generation with composable diffusion models.
\newblock In {\em ECCV}, 2022.

\bibitem{liu2021more}
Xihui Liu, Dong~Huk Park, Samaneh Azadi, Gong Zhang, Arman Chopikyan, Yuxiao
  Hu, Humphrey Shi, Anna Rohrbach, and Trevor Darrell.
\newblock More control for free! image synthesis with semantic diffusion
  guidance.
\newblock {\em arXiv preprint arXiv:2112.05744}, 2021.

\bibitem{liu2020describe}
Yahui Liu, Marco De~Nadai, Deng Cai, Huayang Li, Xavier Alameda-Pineda, Nicu
  Sebe, and Bruno Lepri.
\newblock Describe what to change: A text-guided unsupervised image-to-image
  translation approach.
\newblock In {\em Proceedings of the 28th ACM International Conference on
  Multimedia}, 2020.

\bibitem{liu2015faceattributes}
Ziwei Liu, Ping Luo, Xiaogang Wang, and Xiaoou Tang.
\newblock Deep learning face attributes in the wild.
\newblock In {\em ICCV}, 2015.

\bibitem{lugmayr2022repaint}
Andreas Lugmayr, Martin Danelljan, Andres Romero, Fisher Yu, Radu Timofte, and
  Luc Van~Gool.
\newblock Repaint: Inpainting using denoising diffusion probabilistic models.
\newblock In {\em CVPR}, 2022.

\bibitem{meng2021sdedit}
Chenlin Meng, Yutong He, Yang Song, Jiaming Song, Jiajun Wu, Jun-Yan Zhu, and
  Stefano Ermon.
\newblock Sdedit: Guided image synthesis and editing with stochastic
  differential equations.
\newblock In {\em International Conference on Learning Representations}, 2021.

\bibitem{nichol2021-glide}
Alex Nichol, Prafulla Dhariwal, Aditya Ramesh, Pranav Shyam, Pamela Mishkin,
  Bob McGrew, Ilya Sutskever, and Mark Chen.
\newblock Glide: Towards photorealistic image generation and editing with
  text-guided diffusion models, 2021.

\bibitem{paige2017learning}
Brooks Paige, Jan-Willem van~de Meent, Alban Desmaison, Noah Goodman, Pushmeet
  Kohli, Frank Wood, Philip Torr, et~al.
\newblock Learning disentangled representations with semi-supervised deep
  generative models.
\newblock In {\em Neurips}, 2017.

\bibitem{patashnik2021styleclip}
Or Patashnik, Zongze Wu, Eli Shechtman, Daniel Cohen-Or, and Dani Lischinski.
\newblock Styleclip: Text-driven manipulation of stylegan imagery.
\newblock In {\em ICCV}, 2021.

\bibitem{preechakul2022diffusion}
Konpat Preechakul, Nattanat Chatthee, Suttisak Wizadwongsa, and Supasorn
  Suwajanakorn.
\newblock Diffusion autoencoders: Toward a meaningful and decodable
  representation.
\newblock In {\em CVPR}, 2022.

\bibitem{radford2021learning}
Alec Radford, Jong~Wook Kim, Chris Hallacy, Aditya Ramesh, Gabriel Goh,
  Sandhini Agarwal, Girish Sastry, Amanda Askell, Pamela Mishkin, Jack Clark,
  et~al.
\newblock Learning transferable visual models from natural language
  supervision.
\newblock In {\em ICML}, 2021.

\bibitem{ramesh2022hierarchical}
Aditya Ramesh, Prafulla Dhariwal, Alex Nichol, Casey Chu, and Mark Chen.
\newblock Hierarchical text-conditional image generation with clip latents.
\newblock {\em arXiv preprint arXiv:2204.06125}, 2022.

\bibitem{NEURIPS2019_vq-vae}
Ali Razavi, Aaron van~den Oord, and Oriol Vinyals.
\newblock Generating diverse high-fidelity images with vq-vae-2.
\newblock In {\em Neurips}, 2019.

\bibitem{ren2021learning}
Xuanchi Ren, Tao Yang, Yuwang Wang, and Wenjun Zeng.
\newblock Learning disentangled representation by exploiting pretrained
  generative models: A contrastive learning view.
\newblock In {\em International Conference on Learning Representations}, 2021.

\bibitem{rezende2015variational}
Danilo Rezende and Shakir Mohamed.
\newblock Variational inference with normalizing flows.
\newblock In {\em ICML}, 2015.

\bibitem{pmlr-v32-rezende14}
Danilo~Jimenez Rezende, Shakir Mohamed, and Daan Wierstra.
\newblock Stochastic backpropagation and approximate inference in deep
  generative models.
\newblock In {\em ICML}, 2014.

\bibitem{rombach2022high}
Robin Rombach, Andreas Blattmann, Dominik Lorenz, Patrick Esser, and Bj{\"o}rn
  Ommer.
\newblock High-resolution image synthesis with latent diffusion models.
\newblock In {\em CVPR}, 2022.

\bibitem{ronneberger2015u}
Olaf Ronneberger, Philipp Fischer, and Thomas Brox.
\newblock U-net: Convolutional networks for biomedical image segmentation.
\newblock In {\em International Conference on Medical image computing and
  computer-assisted intervention}, 2015.

\bibitem{ruiz2022dreambooth}
Nataniel Ruiz, Yuanzhen Li, Varun Jampani, Yael Pritch, Michael Rubinstein, and
  Kfir Aberman.
\newblock Dreambooth: Fine tuning text-to-image diffusion models for
  subject-driven generation.
\newblock {\em arXiv preprint arXiv:2208.12242}, 2022.

\bibitem{saharia2022palette}
Chitwan Saharia, William Chan, Huiwen Chang, Chris Lee, Jonathan Ho, Tim
  Salimans, David Fleet, and Mohammad Norouzi.
\newblock Palette: Image-to-image diffusion models.
\newblock In {\em ACM SIGGRAPH 2022 Conference Proceedings}, 2022.

\bibitem{saharia2022photorealistic}
Chitwan Saharia, William Chan, Saurabh Saxena, Lala Li, Jay Whang, Emily
  Denton, Seyed Kamyar~Seyed Ghasemipour, Burcu~Karagol Ayan, S~Sara Mahdavi,
  Rapha~Gontijo Lopes, et~al.
\newblock Photorealistic text-to-image diffusion models with deep language
  understanding.
\newblock {\em arXiv preprint arXiv:2205.11487}, 2022.

\bibitem{saharia2022-image}
Chitwan Saharia, Jonathan Ho, William Chan, Tim Salimans, David~J. Fleet, and
  Mohammad Norouzi.
\newblock Image super-resolution via iterative refinement.
\newblock {\em IEEE Transactions on Pattern Analysis and Machine Intelligence},
  pages 1--14, 2022.

\bibitem{schuhmann2021laion}
Christoph Schuhmann, Richard Vencu, Romain Beaumont, Robert Kaczmarczyk,
  Clayton Mullis, Aarush Katta, Theo Coombes, Jenia Jitsev, and Aran
  Komatsuzaki.
\newblock Laion-400m: Open dataset of clip-filtered 400 million image-text
  pairs.
\newblock {\em arXiv preprint arXiv:2111.02114}, 2021.

\bibitem{shen2020interpreting}
Yujun Shen, Jinjin Gu, Xiaoou Tang, and Bolei Zhou.
\newblock Interpreting the latent space of gans for semantic face editing.
\newblock In {\em CVPR}, 2020.

\bibitem{shen2020interfacegan}
Yujun Shen, Ceyuan Yang, Xiaoou Tang, and Bolei Zhou.
\newblock Interfacegan: Interpreting the disentangled face representation
  learned by gans.
\newblock {\em IEEE transactions on pattern analysis and machine intelligence},
  2020.

\bibitem{shen2021closed}
Yujun Shen and Bolei Zhou.
\newblock Closed-form factorization of latent semantics in gans.
\newblock In {\em CVPR}, 2021.

\bibitem{sohl2015deep}
Jascha Sohl-Dickstein, Eric Weiss, Niru Maheswaranathan, and Surya Ganguli.
\newblock Deep unsupervised learning using nonequilibrium thermodynamics.
\newblock In {\em International Conference on Machine Learning}. PMLR, 2015.

\bibitem{song2020denoising}
Jiaming Song, Chenlin Meng, and Stefano Ermon.
\newblock Denoising diffusion implicit models.
\newblock {\em arXiv preprint arXiv:2010.02502}, 2020.

\bibitem{song2020score}
Yang Song, Jascha Sohl-Dickstein, Diederik~P Kingma, Abhishek Kumar, Stefano
  Ermon, and Ben Poole.
\newblock Score-based generative modeling through stochastic differential
  equations.
\newblock In {\em ICLR}, 2021.

\bibitem{tashiro2021-csdi}
Yusuke Tashiro, Jiaming Song, Yang Song, and Stefano Ermon.
\newblock Csdi: Conditional score-based diffusion models for probabilistic time
  series imputation, 2021.

\bibitem{wang2022unifying}
Peng Wang, An Yang, Rui Men, Junyang Lin, Shuai Bai, Zhikang Li, Jianxin Ma,
  Chang Zhou, Jingren Zhou, and Hongxia Yang.
\newblock Unifying architectures, tasks, and modalities through a simple
  sequence-to-sequence learning framework.
\newblock In {\em ICML}, 2022.

\bibitem{yu2015lsun}
Fisher Yu, Ari Seff, Yinda Zhang, Shuran Song, Thomas Funkhouser, and Jianxiong
  Xiao.
\newblock Lsun: Construction of a large-scale image dataset using deep learning
  with humans in the loop.
\newblock {\em arXiv preprint arXiv:1506.03365}, 2015.

\bibitem{zhu2017unpaired}
Jun-Yan Zhu, Taesung Park, Phillip Isola, and Alexei~A Efros.
\newblock Unpaired image-to-image translation using cycle-consistent
  adversarial networks.
\newblock In {\em Proceedings of the IEEE international conference on computer
  vision}, 2017.

\end{thebibliography}
}

\newpage
\newpage
\clearpage
\maketitle
\begin{table*}
    \centering
    \begin{tabular}{lll}
    \toprule
     &  & Value \\
    \midrule
    \multirow{8}{*}{\textbf{Optimization}} & \multirow{2}{*}{\e{\beta}}  & 0.05 (attributes on person)\\
        & & 0.03 (attributes on scenes)\\
        & Optimizer & Adam~\cite{kingma2014adam} \\
        & Learning rate & 0.05\\
        & \multirow{2}{*}{\e{\lambda_t} initialization} & $0.0$ for $t\geq 0.8T$, $1.0$ for $t<0.8T$ (attributes on person) \\
        & & $0.0$ for $t\geq 0.9T$, $1.2$ for $t<0.9T$ (attributes on scenes)\\
        & Checkpoint for CLIP loss &  \texttt{ViT-B/32}~\cite{ren2021learning} \\
        & Checkpoint for perceptual loss & \texttt{VGG-16}~\cite{johnson2016perceptual} \\
    \midrule
    \multirow{7}{*}{\textbf{Diffusion Model}} & Model checkpoint  & \texttt{stable-diffusion-v1-4}~\cite{ackermann2022high} \\
        & Sampling steps & 50\\
        & Sampling variance & 0.0\\
        & Resolution & $512\times 512$\\
        & Latent channels & 4 \\
        & Latent down-sampling factor & 8 \\
        & Conditional guidance scale & 7.5 \\
    \bottomrule
    \end{tabular}
    \caption{\textbf{Hyperparameter settings and model architectures used in this paper.}}
    \label{tab:supp-hyperparameters}
\end{table*}

\begin{alphasection}
\section{Implementation Details}\label{hyperparameters}
To help reproduce our results, we report the detailed hyperparameter settings and model architectures used in this work in Table~\ref{tab:supp-hyperparameters}.

In terms of optimization, we use different loss balancing weight \e{\beta} and different initialization of combination weights \e{\lambda_t} for attributes on person and scenes, since edits on person require more strict content preservation than edits on scenes. Specifically, when editing person, we adopt a larger \e{\beta} on perceptual loss and initialize \e{\lambda_t} such that \e{\bm c_{1:T}} is more similar to the style-neutral description \e{\bm c^{(0)}}.
Both these settings encourage content preservation when disentangling attributes on person.
For the directional CLIP loss, we use the pre-trained \texttt{ViT-B/32}~\cite{ren2021learning}. For the model used to compute perceptual loss, we adopt pre-trained \texttt{VGG-16}~\cite{johnson2016perceptual}.
The stable diffusion model we use is the pre-trained \texttt{stable-diffusion-v1-4}~\cite{ackermann2022high}.
We use all pre-trained models without changing any parameters.

\section{Text Descriptions Used for Attribute Disentanglement and Image Editing}\label{text-list}
In this section, we provide the exact text descriptions we use to disentangle target attributes and perform image editing in this paper. For each target attribute or edit, the text description consists of a style-neutral description and a description with explicit styles, whose embeddings are denoted as \e{\bm c^{(0)}} and \e{\bm c^{(1)}} respectively. For brevity, we use these notations to represent their corresponding text descriptions and list them in Table~\ref{tab:text_description}. For each attribute in the table, we also provide the corresponding figure that demonstrates the visual effects.

We emphasize that our method is generally robust to the choice of text descriptions and is not restricted to the text listed here. Please refer to Sec.~\ref{ablation-main} and Sec.~\ref{text-influence} for more analyses on the robustness of our method to different choices of text descriptions.

\begin{table*}
    \centering
    \resizebox{1.0\linewidth}{!}{%
    \begin{tabular}{lllll}
    \toprule
    Type & Attribute & \e{\bm c^{(0)}} & \e{\bm c^{(1)}} & Example \\
    \midrule
    \multirow{15}{*}{\textbf{Scenes}} & \textbf{Global attributs:} & & & \\ & Children Drawing  & A castle & A castle, children drawing style & Fig.~\ref{fig:exp1} \\
         &  Cyberpunk Style & A street view & A street view, Cyberpunk style & Fig.~\ref{fig:exp1}\\
         & Anime Style & A lake in mountains & A lake in mountains, anime style & Fig.~\ref{fig:supp-exp-G_1}\\
         & Wooden Building & A photo of church exterior & A photo of church exterior, wooden style & Fig.~\ref{fig:supp-subjective-lsun}\\
         & Golden Building & A photo of church exterior & A photo of church exterior, golden style & Fig.~\ref{fig:exp1}\\
         & Red Brick Building & A photo of church exterior & A photo of church exterior, red brick & Fig.~\ref{fig:intro} \\
         & In Sunset & A photo of church exterior & A photo of church exterior, in sunset & Fig.~\ref{fig:supp-subjective-lsun} \\
         & At Dark Night & A photo of seaside & A photo of seaside, dark night & Fig.~\ref{fig:qualitative}\\
         & At Starry Night & A photo of railway & A photo of railway, in milky galaxy & Fig.~\ref{fig:supp-more-disentangle}\\
         & Covered by Snow &  A photo of church exterior & A photo of church exterior, covered by snow & Fig.~\ref{fig:supp-exp-G_1}\\
         & \textbf{Local attributes:} & & & \\
         & Cherry Blossom & A forest & A forest, cherry blossom & Fig.~\ref{fig:exp1}\\
         & Rainbow & A lake in mountains & A lake in mountains, rainbow & Fig.~\ref{fig:qualitative} \\
         & Foothills & A man sitting on grass & A man sitting on grass, in mountains & Fig.~\ref{fig:qualitative}\\
     \midrule
     \multirow{12}{*}{\textbf{Person}} & \textbf{Global attributes:} & & & \\ 
     & Renaissance Style  & A photo of person & A photo of person, renaissance style & Fig.~\ref{fig:intro} \\
         & Egyptian Mural Style & A photo of person & A photo of person, Egyptian mural style & Fig.~\ref{fig:exp1} \\
         & Sketch & A photo of person & A photo of person, sketch style & Fig.~\ref{fig:supp-subjective-celeba} \\
         & Pixar & A photo of person & A photo of person, pixar style & Fig.~\ref{fig:supp-subjective-celeba}\\
         & Young & A photo of person & A photo of person, young & Fig.~\ref{fig:exp1}\\
         & Tanned & A photo of face & A photo of face, tanned & Fig.~\ref{fig:supp-subjective-celeba} \\
         & Male & A photo of face & A photo of face, male & Fig.~\ref{fig:supp-subjective-celeba}\\
         & \textbf{Local attributes:} & & & \\
         & Smiling & A photo of person & A photo of person, smiling & Fig.~\ref{fig:supp-exp-G_1}\\
         & Crying & A photo of person & A photo of person, crying & Fig.~\ref{fig:supp-more-disentangle} \\
         & Angry & A photo of person & A photo of person, angry & Fig.~\ref{fig:supp-more-disentangle} \\         
    \bottomrule
    \end{tabular}
    }
    \caption{\textbf{Text descriptions used for attributes disentanglement and image editing in this paper.} For each attribute, we report the descriptions used for \e{\bm c^{(0)}} and \e{\bm c^{(1)}} as well as the corresponding visual example in the paper.}
    \label{tab:text_description}
\end{table*}

\section{Additional Examples on Partially Replacing Text Embeddings}\label{3.2disentangle}
As described in Sec.\ref{sec3.2}, the stable diffusion model is inherently capable of disentangling attributes, and we demonstrate that such disentanglement can be triggered by partially replacing the text embeddings from a style-neutral one to the one with explicit styles. In this section, we provide more examples to better illustrate this phenomenon.

The examples are shown in Fig.~\ref{fig:supp-exp-C}. In the figure, each row demonstrates an example of replacing the text embedding at later denoising steps.
In other words,  we use the style-neutral description \e{\bm c^{(0)}} during early denoising steps (\e{T} to \e{t'}), and replace it with the one containing explicit styles \e{\bm c^{(1)}} during later steps (\e{t'} to 0).
\e{\bm c^{(0)}} and \e{\bm c^{(1)}} are listed on the left and right of each row, respectively. In the first column, \e{t'} is set to 0, which corresponds to using \e{\bm c^{(0)}} for all denoising steps. On the other hand, \e{t'=T} in the last column means the text embedding replacement happens at the beginning, and the denoising is entirely conditioned on \e{\bm c^{(1)}}.

From these results, we observe that only replacing \e{\bm c^{(0)}} with \e{\bm c^{(1)}} in later denoising steps can maintain the contents in the style-neutral image, and more replaced steps lead to stronger modification effects on the target attribute. Thus for some specific time steps \e{t'} (\emph{e.g.,} \e{t'=0.8T} for ``red brick'' and \e{t'=0.9T} for ''renaissance style''), the target attribute can be successfully disentangled.
This verifies the inherent disentanglement ability in the stable diffusion model. 
Furthermore, we 
observe that for different attributes, the optimal \e{t'} could be different. For example, \e{t'=0.8T} disentangles the ``red brick'' and ``in sunset'' attributes but does not bring successful modifications for other two attributes.
In other words, one has to search the best \e{t'} for optimal disentanglement.
This motivates a more principled optimization scheme to combine text embeddings for the best disentanglement, which is described in Sec.~\ref{optimization}.

\begin{figure*}
\centering
\includegraphics[width=0.97\textwidth]{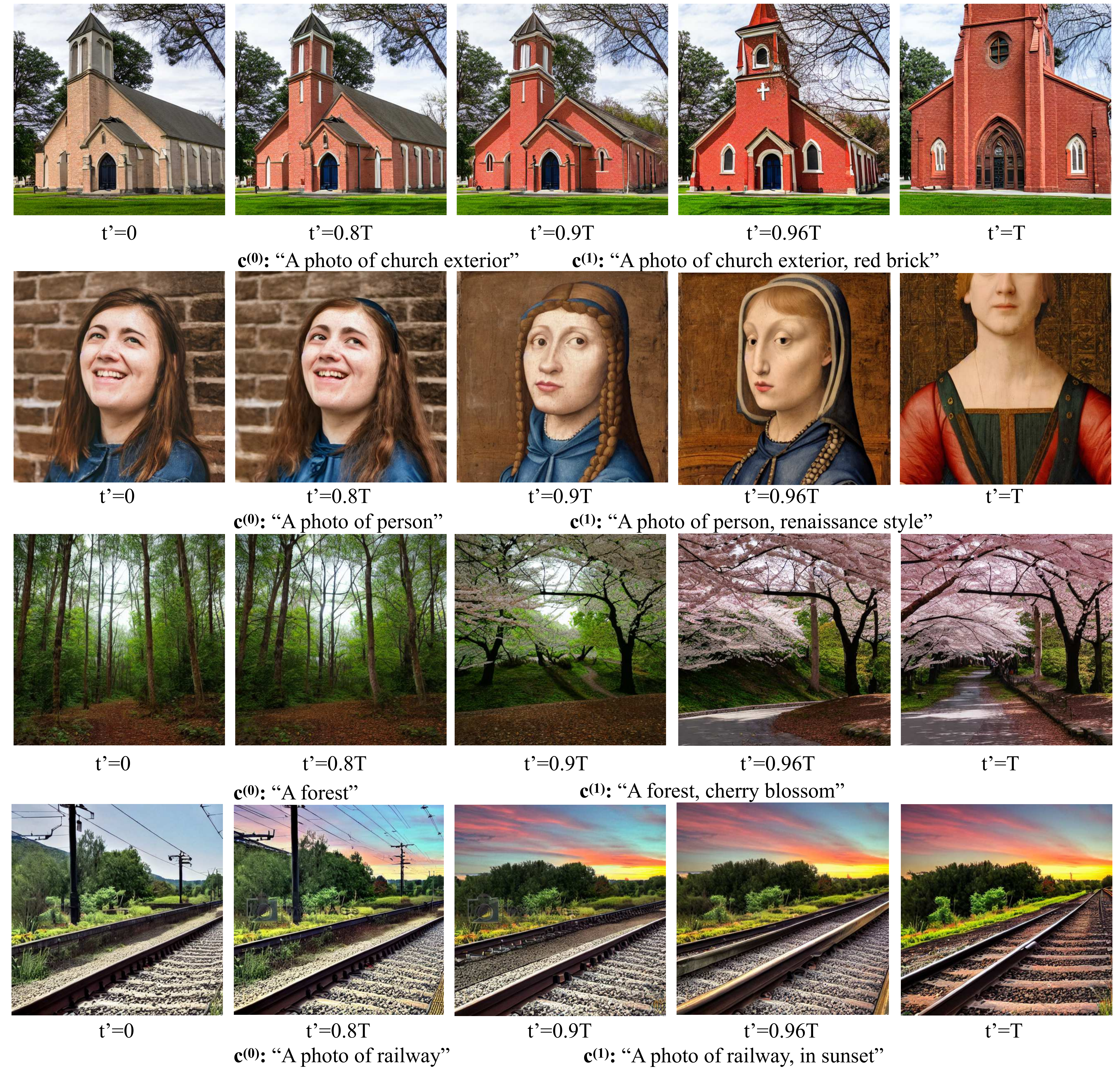}
\caption{\textbf{Inherent disentanglement capability in the stable diffusion model.}
For each row, we partially replace the style-neutral text description \e{\bm c^{(0)}} with another description \e{\bm c^{(1)}} that includes the explicit style. Particularly, the denoising process conditions on \e{\bm c^{(0)}} from \e{T} to \e{t'} and \e{\bm c^{(1)}} from \e{t'} to 0.}
\vspace*{-0.1in}
\label{fig:supp-exp-C}
\end{figure*}

\section{More Examples of Disentangled Attributes}\label{4.1more_example}
We now provide more examples of attributes that can be disentangled by our method in Fig.~\ref{fig:supp-more-disentangle}. These results show that our method is generalizable to a broad range of attributes, and it satisfies the two criteria for disentanglement discussed in Sec.~\ref{sec:intro}.
\begin{figure*}
\centering
\vspace{-0.5em}
\includegraphics[width=0.98\textwidth]{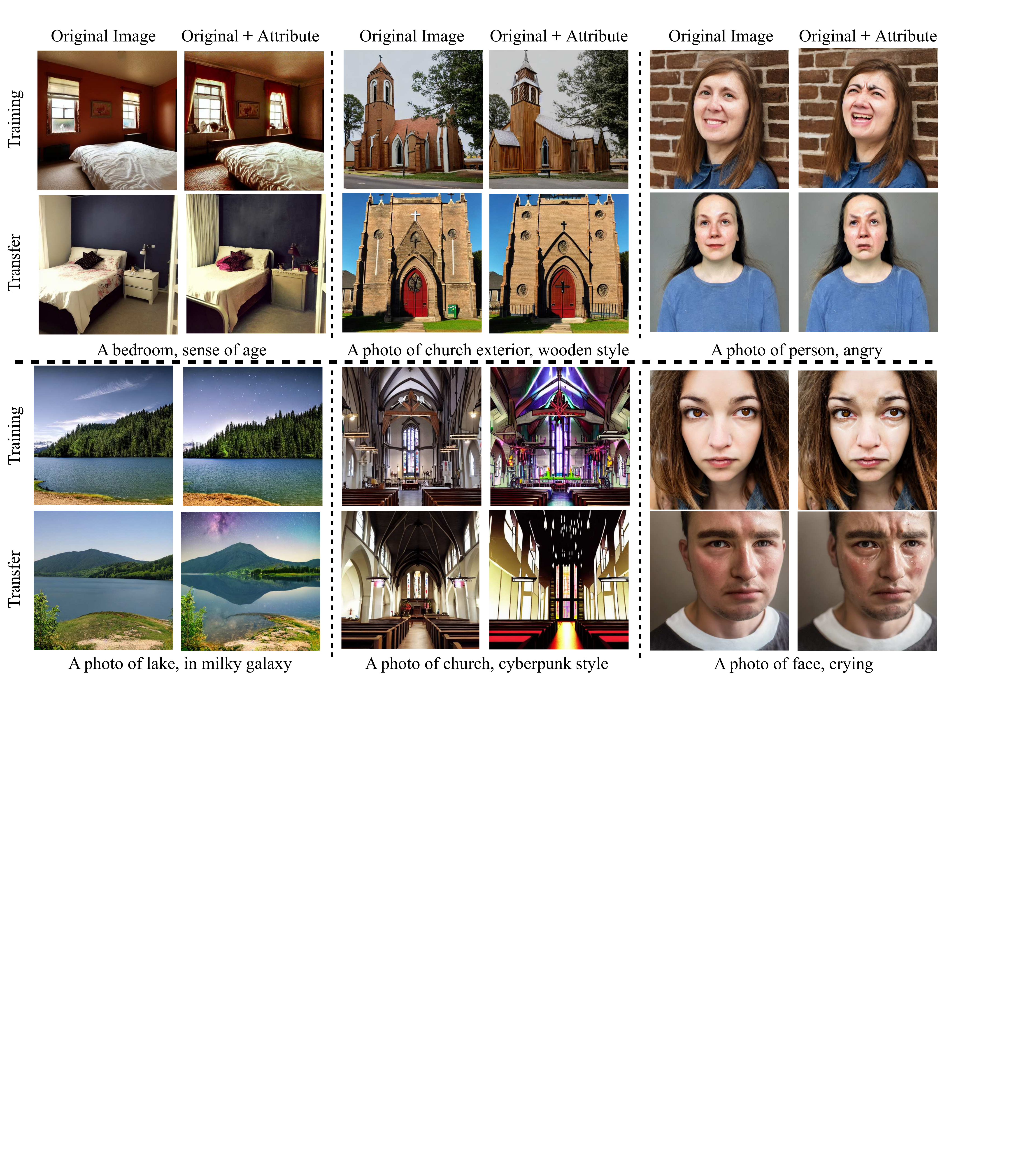}
\vspace{-1em}
\caption{\textbf{Examples of disentangled attributes}. Text description with style (\e{\bm c^{(1)}}) is shown below each row, which consists of the style-neutral text description (\e{\bm c^{(0)}}) and target attribute description, separated by comma. Within each attribute, \textbf{first row:} results on optimization images; \textbf{second row:} results of transferring to unseen images; \textbf{left column:} source images; \textbf{right column:} modified images.}
\vspace*{-0.1in}
\label{fig:supp-more-disentangle}
\end{figure*}

\section{Details of Subjective Evaluation}\label{human_study}
In this section, we detail our subjective evaluation process.
We compare the performance of our method with \textsc{DiffusionCLIP}~\cite{kim2022diffusionclip} on the image editing task.
Specifically, we consider two existing datasets used by \textsc{DiffusionCLIP}:
Celeb-A~\cite{sohl2015deep} that focuses on person and LSUN-church~\cite{yu2015lsun} that focuses on churches.
For each dataset, we select the first 20 images as source images\footnote{For the ``male'' attribute, we use 20 female images from Celeb-A dataset as source images.} and evaluate 4 types of edits used in \cite{kim2022diffusionclip}, \emph{i.e.,} tanned, male, sketch, and pixar for human faces, and golden, wooden, snowy, and red brick for churches.
We compare our editing results with the results generated by the official checkpoints of \textsc{DiffusionCLIP} that are fine-tuned for each target edit.
We conduct evaluation on Amazon Mechanical Turk, and we require all participants to be master workers in order to answer our questions. In total, 11 workers participate in the study. 
For each edit, we present participants with the source image, as well as images edited by two methods in random order. We ask participants to answer the following three questions:
\begin{itemize}[noitemsep]
    \item[(1)] Which one is perceptually consistent with the target edit attribute and looks like a natural image?
    \item[(2)] Which one better preserves the information of original images (e.g. background, shape)?
    \item[(3)] Overall, which editing result is better?
\end{itemize}

Fig.~\ref{fig:human-study} shows the results of subjective evaluation. We also provide all generated images by both methods in Fig.~\ref{fig:supp-subjective-celeba} and Fig.~\ref{fig:supp-subjective-lsun}. We observe that \textsc{DiffusionCLIP} tends to over-change the attribute in the image to the extent that introduces artifacts in the image and modifies other contents (\emph{e.g.,} when making the church golden, it changes the sky and ground into gold). By contrast, our method generates more natural images and only modifies the target object.

\begin{figure*}
\centering
\vspace{-1.0em}
\includegraphics[width=0.83\textwidth]{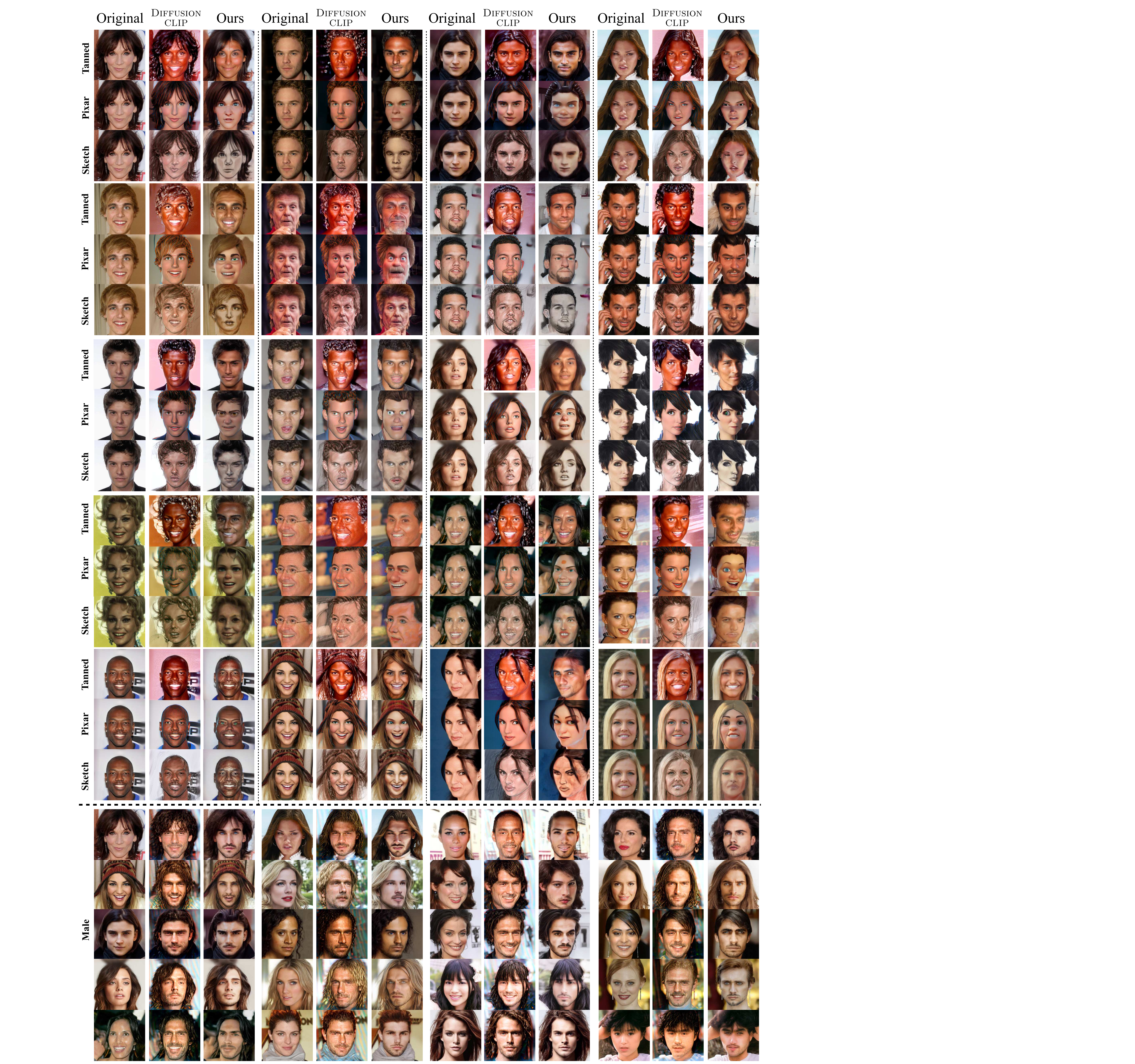}
\vspace{-1em}
\caption{\textbf{Generated images for subjective evaluation on Celeb-A dataset.}
Different from other attributes, 20 female images are used as source images for the attribute ``male''.}
\vspace*{-0.1in}
\label{fig:supp-subjective-celeba}
\end{figure*}

\begin{figure*}
\centering
\vspace{-1.0em}
\includegraphics[width=0.83\textwidth]{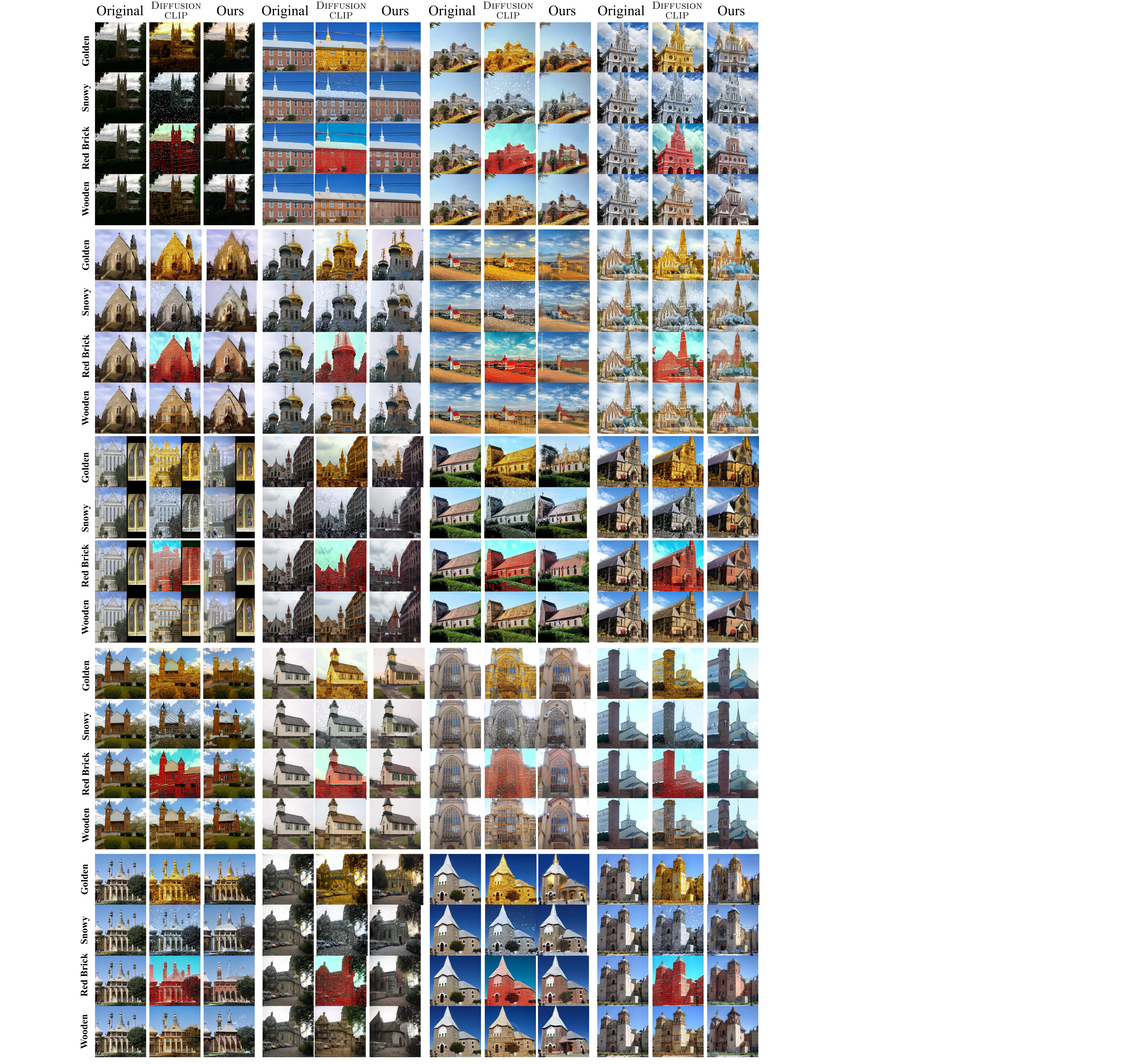}
\vspace{-1em}
\caption{\textbf{Generated images for subjective evaluation on LSUN-church dataset.}
}
\vspace*{-0.1in}
\label{fig:supp-subjective-lsun}
\end{figure*}

\section{More Comparisons with Baselines}\label{compare_baseline_qualitative}

In this section, we perform more qualitative comparison between our method and the state-of-the-art diffusion-model-based image editing methods. As mentioned in Sec.~\ref{human-study}, since these methods either have not released code by the time of our submission or require auxiliary labels that are unavailable, we cannot include them in the subjective evaluation.
Instead, we collect the source and edited images in their papers and perform the same edit using our method for comparison. The results are shown in Fig.~\ref{fig:supp-baseline-compare}. Overall, our method achieves comparable editing results with the baselines. More specifically, our method produces stronger and more natural editing results for global target attributes (\emph{e.g.,} rainy, snowy), while our method has difficulties disentangling attributes for small edits such as cake decorations.
Meanwhile, we comment that the results of comparing with \textsc{DiffusionCLIP} are blurry due to the low-resolution inputs obtained from the original paper. Please refer to Fig.~\ref{fig:supp-subjective-celeba} and Fig.~\ref{fig:supp-subjective-lsun} for a more comprehensive comparison with \textsc{DiffusionCLIP}.

\begin{figure*}
\centering
\vspace{-0.5em}
\includegraphics[width=0.98\textwidth]{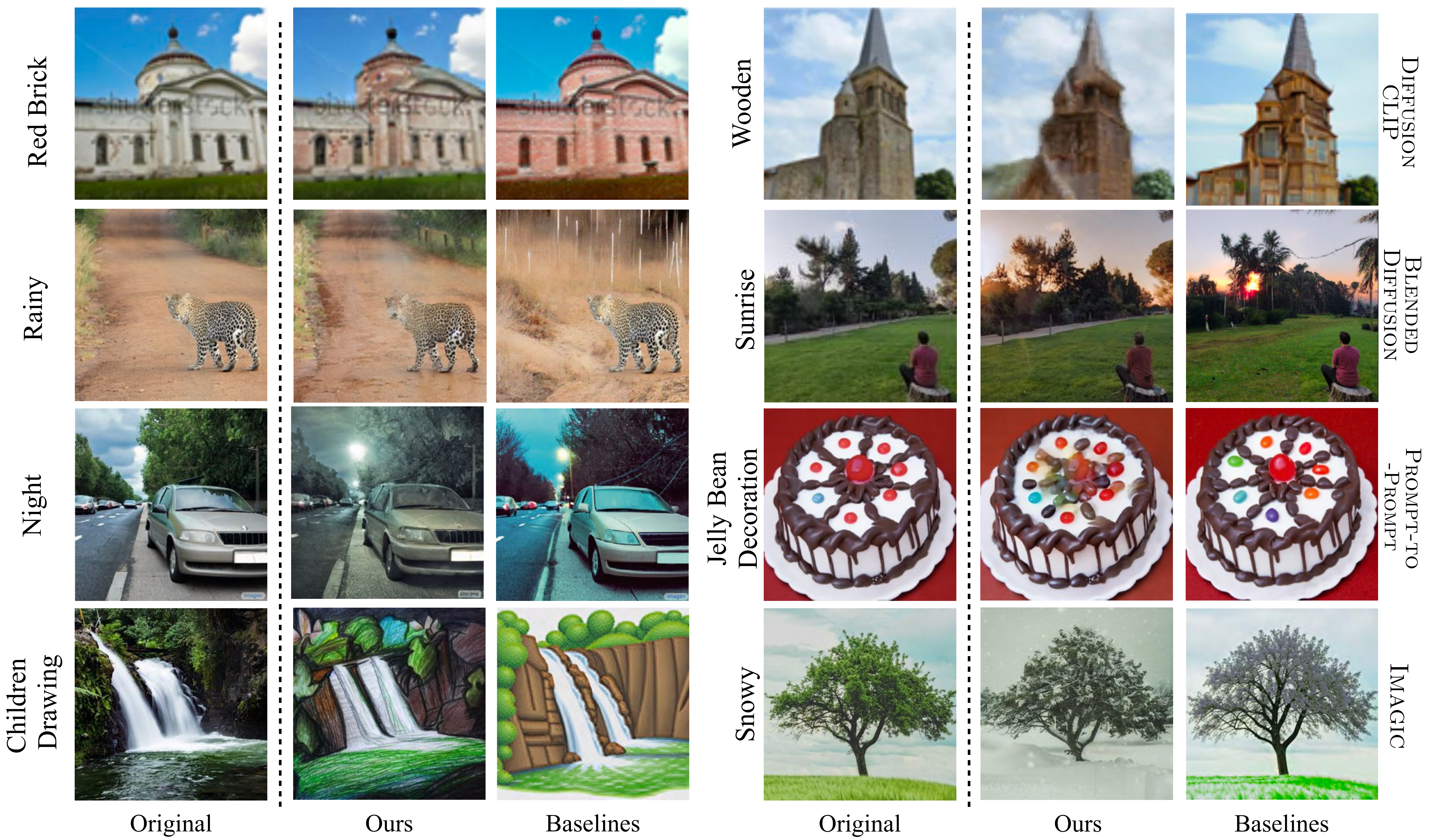}
\vspace{-1em}
\caption{\textbf{Comparisons with baselines}. The source image and corresponding baseline result are taken from the original papers.
Each row illustrates the comparison with one particular baseline.}
\vspace*{-0.1in}
\label{fig:supp-baseline-compare}
\end{figure*}

\section{Effects of Varying Text Descriptions}\label{text-influence}
Here we provide more analyses on whether our method is robust to different choices of text descriptions. We consider three types of variations in the following section.

\vspace*{0.07in}
\noindent \textbf{The way of appending target attributes in \e{\bm c^{(1)}}:} We explore how the way that \e{\bm c^{(1)}} appends the target attribute descriptions can influence the results. Concretely, we fix the style-neutral text description \e{\bm c^{(0)}} and explore three ways of concatenating the target attribute description in \e{\bm c^{(1)}}: direct concatenation, concatenation separated by a comma, and concatenation separated by proper prepositions (``with'' and ``in'').
As can be observed in Fig.~\ref{fig:supp-exp-G_1},
for all three variations, our method consistently produces images with desired attributes on people, buildings, and natural scenes. We comment that this is not meant to be an exhaustive list of all the concatenation ways, but generally, our method is robust to how the target attribute descriptions are appended in \e{\bm c^{(1)}}.

\vspace*{0.07in}
\noindent \textbf{More complex target attribute description in \e{\bm c^{(1)}}:}
We further investigate how would the complexity of the target attribute description in \e{\bm c^{(1)}} affect the image editing results. In this experiment,
we again fix \e{\bm c^{(0)}} but gradually increase the complexity of target attribute descriptions by adding more correlated modifiers (\emph{e.g.,} pink flower, pink tree). As shown in Fig.~\ref{fig:supp-exp-G_2}, using one modifier (the second column) is sufficient for successful edits. Meanwhile, we are able to achieve stronger editing effects with more correlated modifiers. For example, for the ``cherry blossom'' attribute, as we increase the number of correlated modifiers, the flowers become more colorful and bright.

\begin{figure*}
\centering
\vspace{-0.5em}
\includegraphics[width=0.9\textwidth]{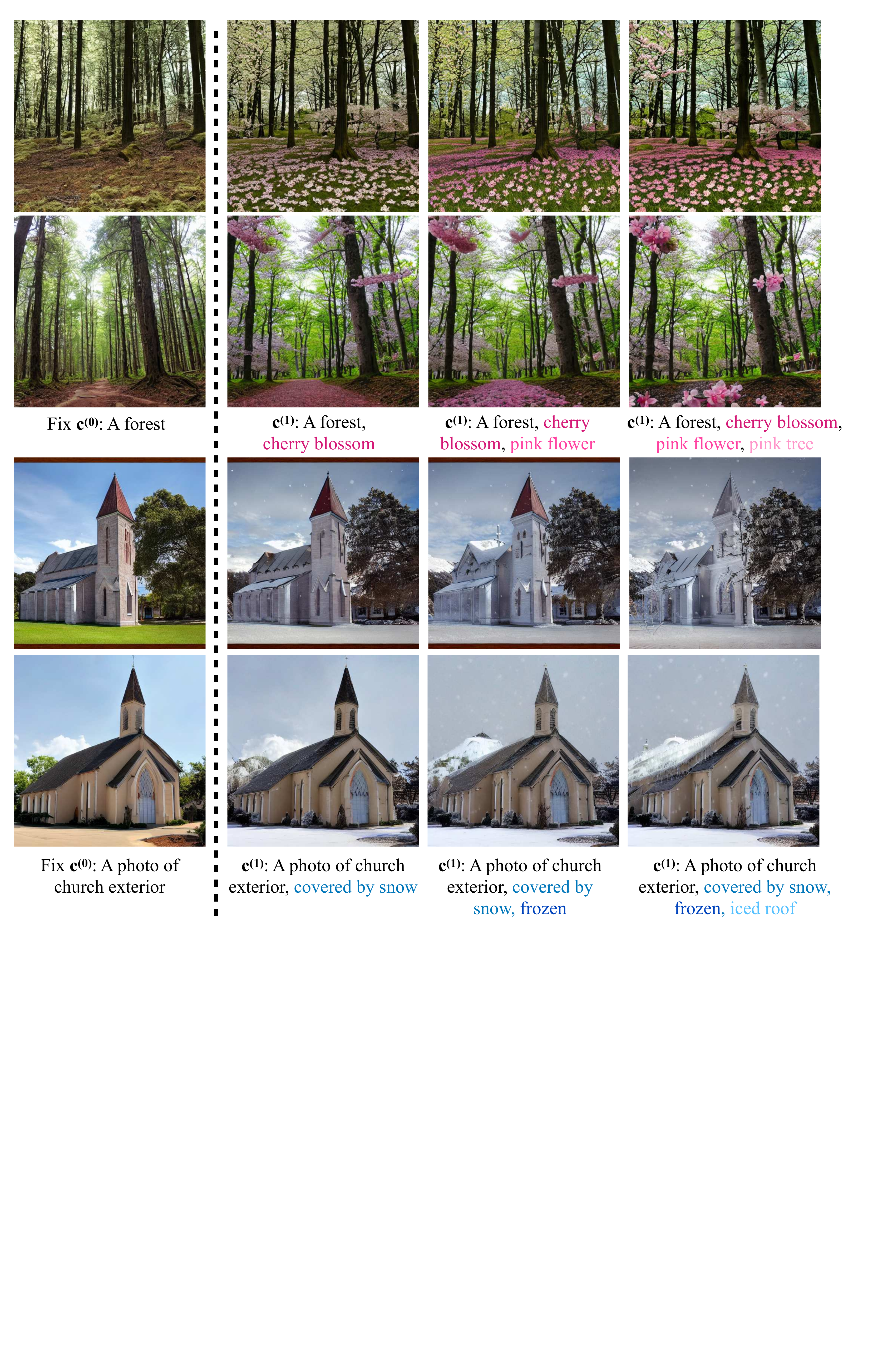}
\vspace{-1.8em}
\caption{\textbf{Effects of changing the complexity of target attribute descriptions in \e{\bm c^{(1)}}.} In each row, we fix \e{\bm c^{(0)}} and increase the complexity of attribute description by adding more correlated modifiers in \e{\bm c^{(1)}}.}
\vspace*{-0.1in}
\label{fig:supp-exp-G_2}
\end{figure*}

\vspace*{0.07in}
\noindent \textbf{The variations of \e{\bm c^{(0)}}:} Lastly, we examine the effects of varying \e{\bm c^{(0)}}. 
We fix the way of appending target attribute descriptions (concatenated by comma) in \e{\bm c^{(1)}}, 
and we examine three variants of \e{\bm c^{(0)}}, including a short description, a longer description by adding non-informative words,
and a description generated by the state-of-the-art image captioning model~\cite{wang2022unifying}.
Results are shown in Fig.~\ref{fig:supp-exp-G_3}. To better compare the effects of variations on \e{\bm c^{(0)}} and \e{\bm c^{(1)}}, we use the same input images and target attributes as in Fig.~\ref{fig:supp-exp-G_1}.
From this figure, we observe the image editing results are largely robust to different
choices of \e{\bm c^{(0)}}, except that
when using the outputs from the captioning model, the editing effects are sometimes not significant. For example, in the last row, the anime style is not shown in the last image.
One possible reason of failure in this case is that the generated caption is long and contains many details, which outweigh the target attribute description.

\begin{figure*}
\centering
\vspace{-0.5em}
\includegraphics[width=0.8\textwidth]{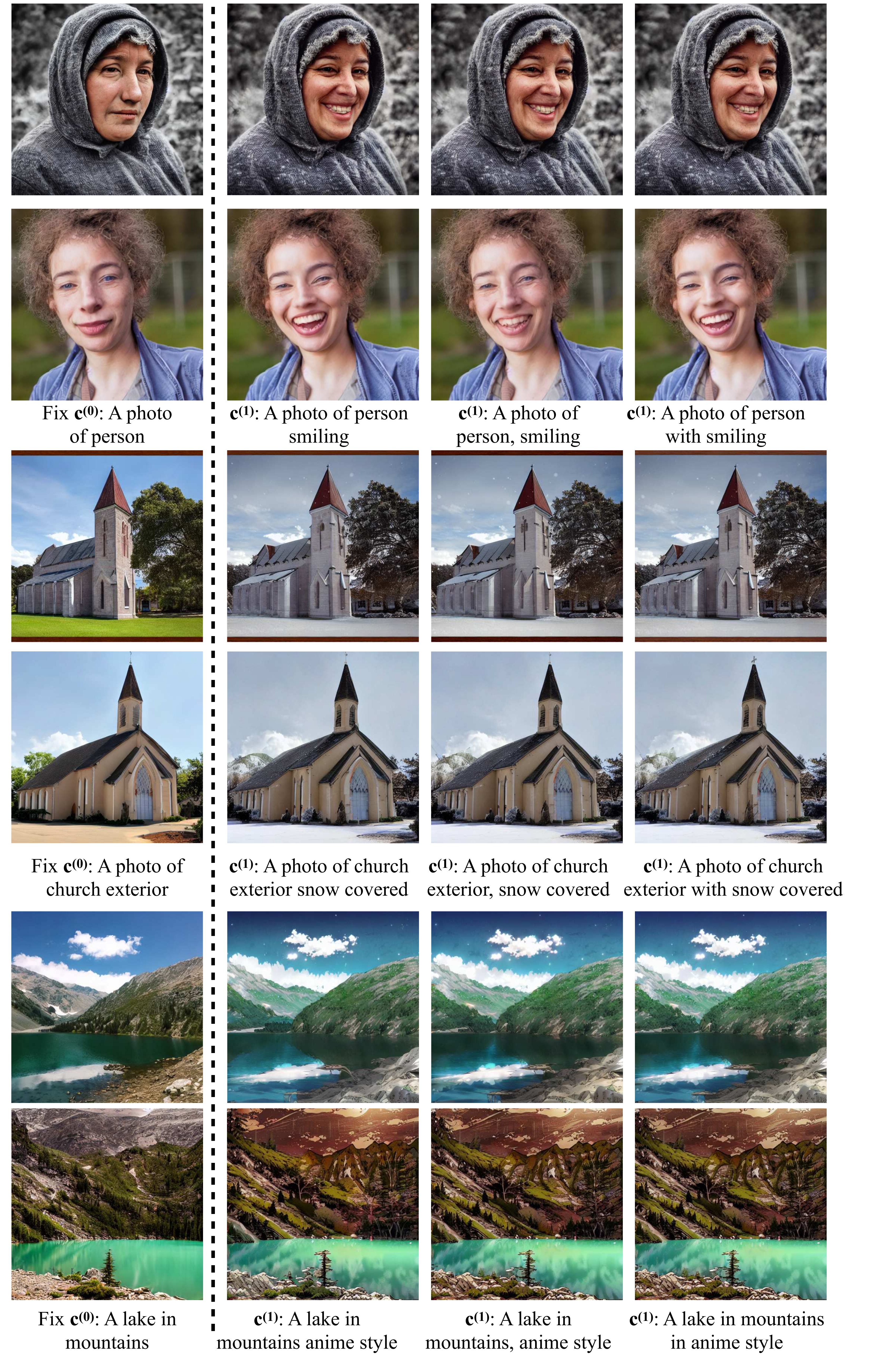}
\vspace{-1em}
\caption{\textbf{Effects of varying the way that \e{\bm c^{(1)}} appends the target attribute description.} In each row, we fix \e{\bm c^{(0)}} and change the way of concatenating target attribute descriptions in \e{\bm c^{(1)}}.}
\vspace*{-0.1in}
\label{fig:supp-exp-G_1}
\end{figure*}

\begin{figure*}
\centering
\vspace{-0.5em}
\includegraphics[width=0.85\textwidth]{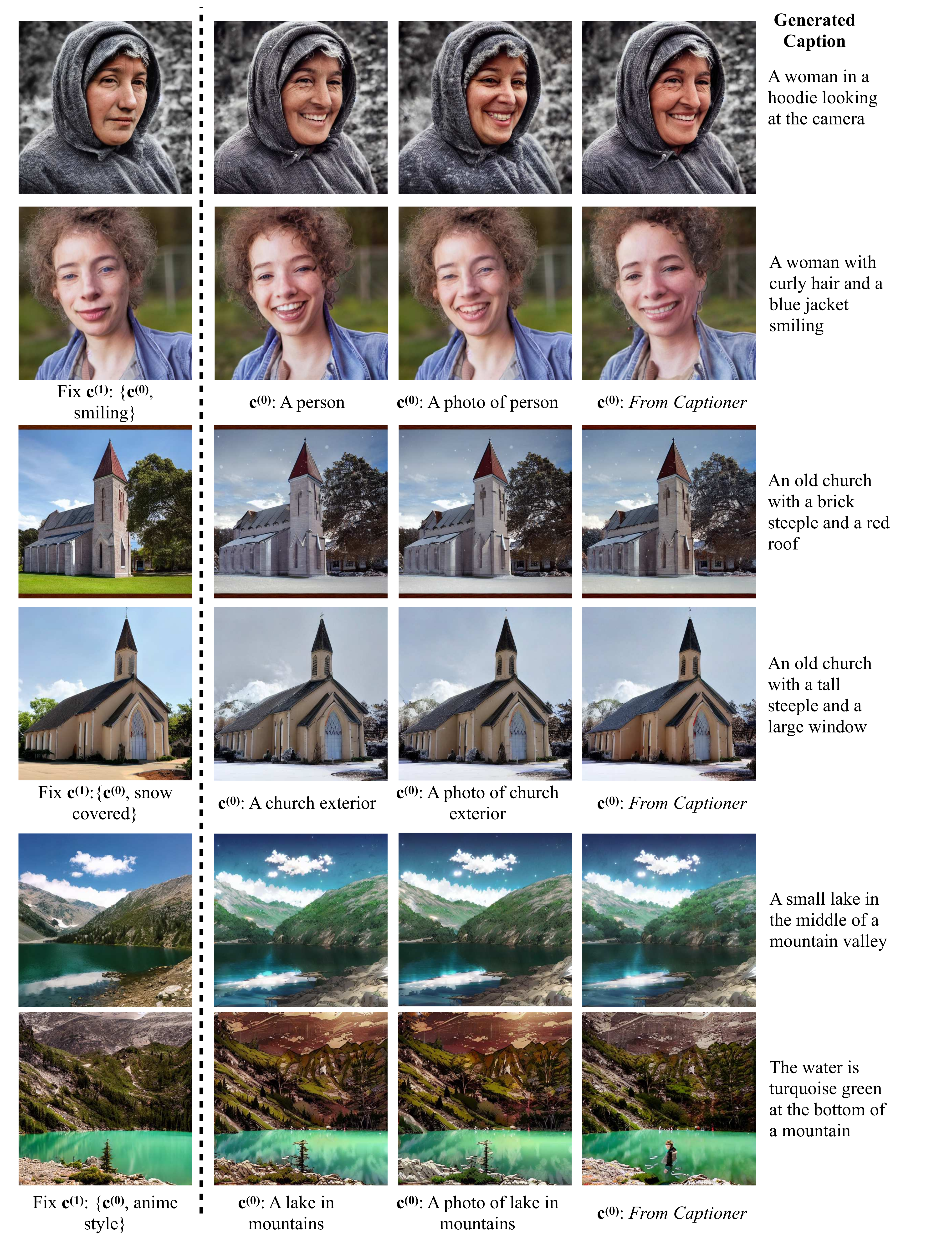}
\vspace{-1em}
\caption{\textbf{Effects of choosing different style-neutral descriptions \e{\bm c^{(0)}}.}
For each row, we fix the target attribute description and consider three variations of \e{\bm c^{(0)}} that describe the same object: a short description, a longer description by adding non-informative words, and a description generated by image captioning model.
}
\vspace*{-0.1in}
\label{fig:supp-exp-G_3}
\end{figure*}

\begin{figure*}
\centering
\includegraphics[width=0.76\textwidth]{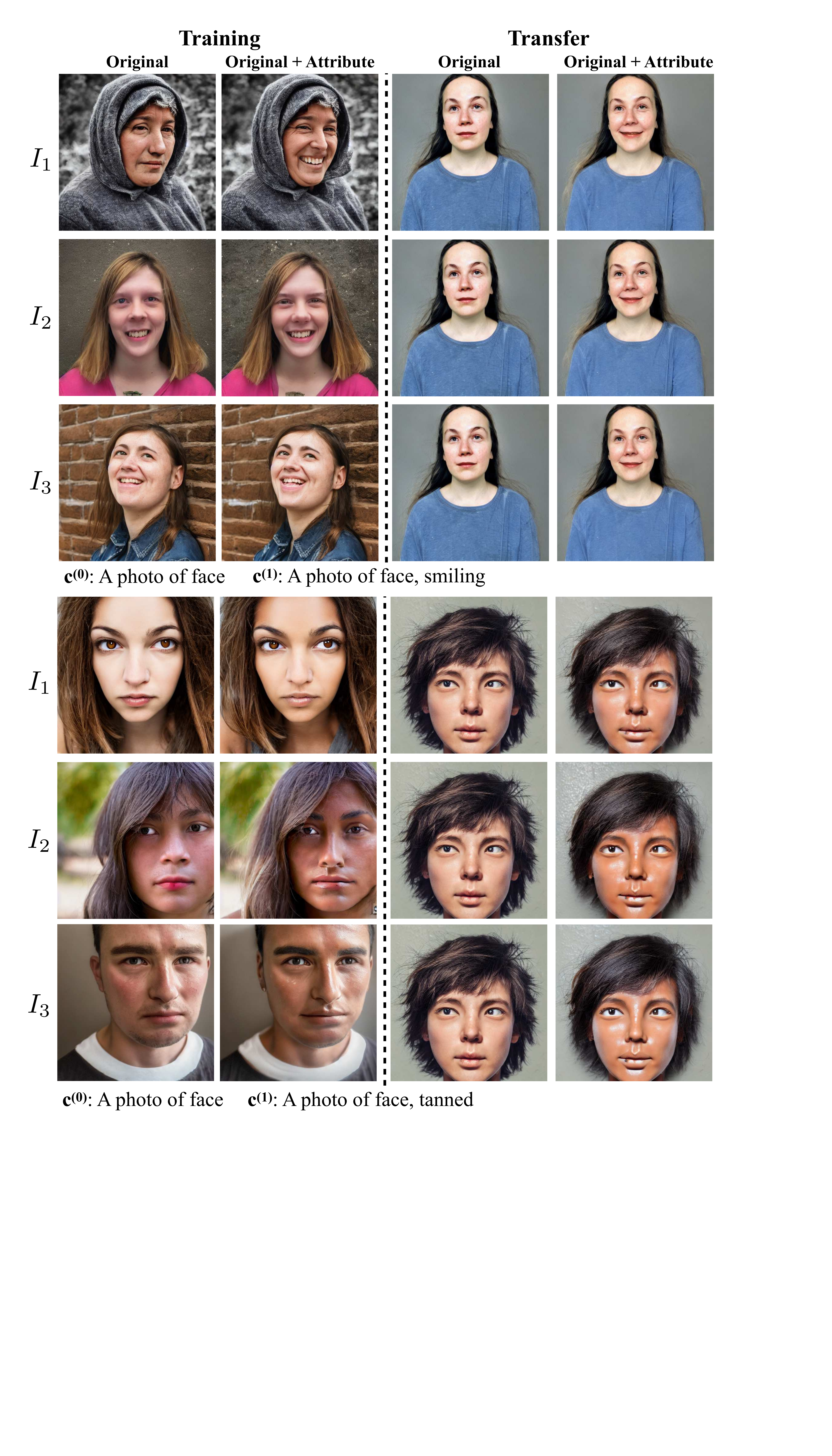}
\caption{
\textbf{Results of using different optimization images.}
For each attribute, \e{I_1, I_2, I_3} represent 3 different optimization images. 
\textbf{Left column:} images used for training and their corresponding disentanglement results. \textbf{Right column:} disentanglement results of applying the learned weights to unseen images.
}
\vspace*{-0.1in}
\label{fig:supp-exp-H}
\end{figure*}

\begin{figure*}
\centering
\includegraphics[width=0.9\textwidth]{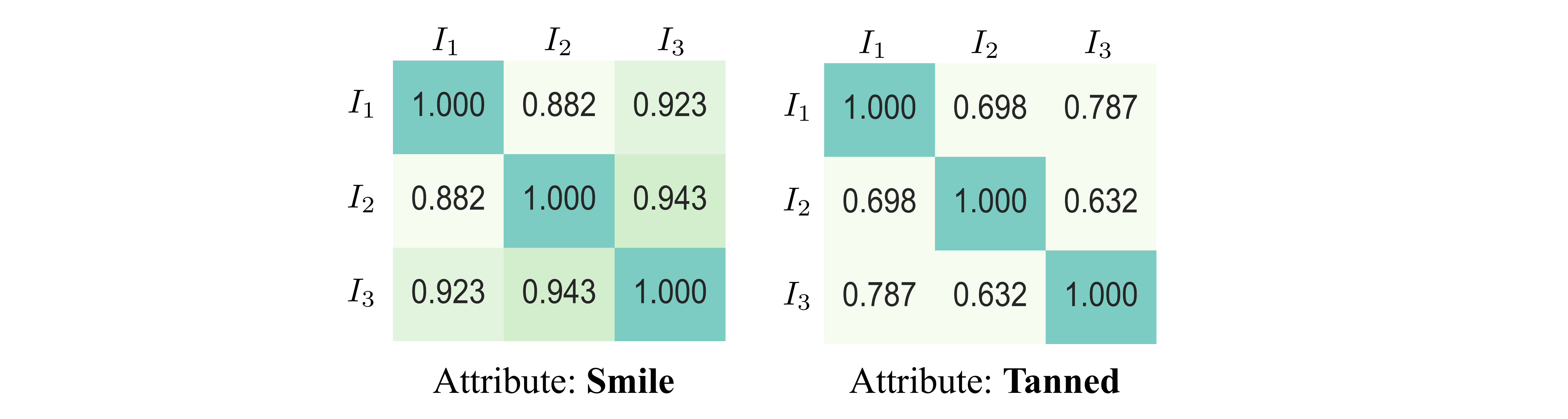}
\vspace{-1em}
\caption{\textbf{Cosine similarity between combination weights \e{\lambda_{1:T}} optimized on different images.}
For each attribute,
\e{I_1, I_2, I_3} represent 3 different optimization images.}
\vspace{-1em}
\label{fig:6.4.1}
\end{figure*}

\begin{figure*}
\centering
\includegraphics[width=0.76\textwidth]{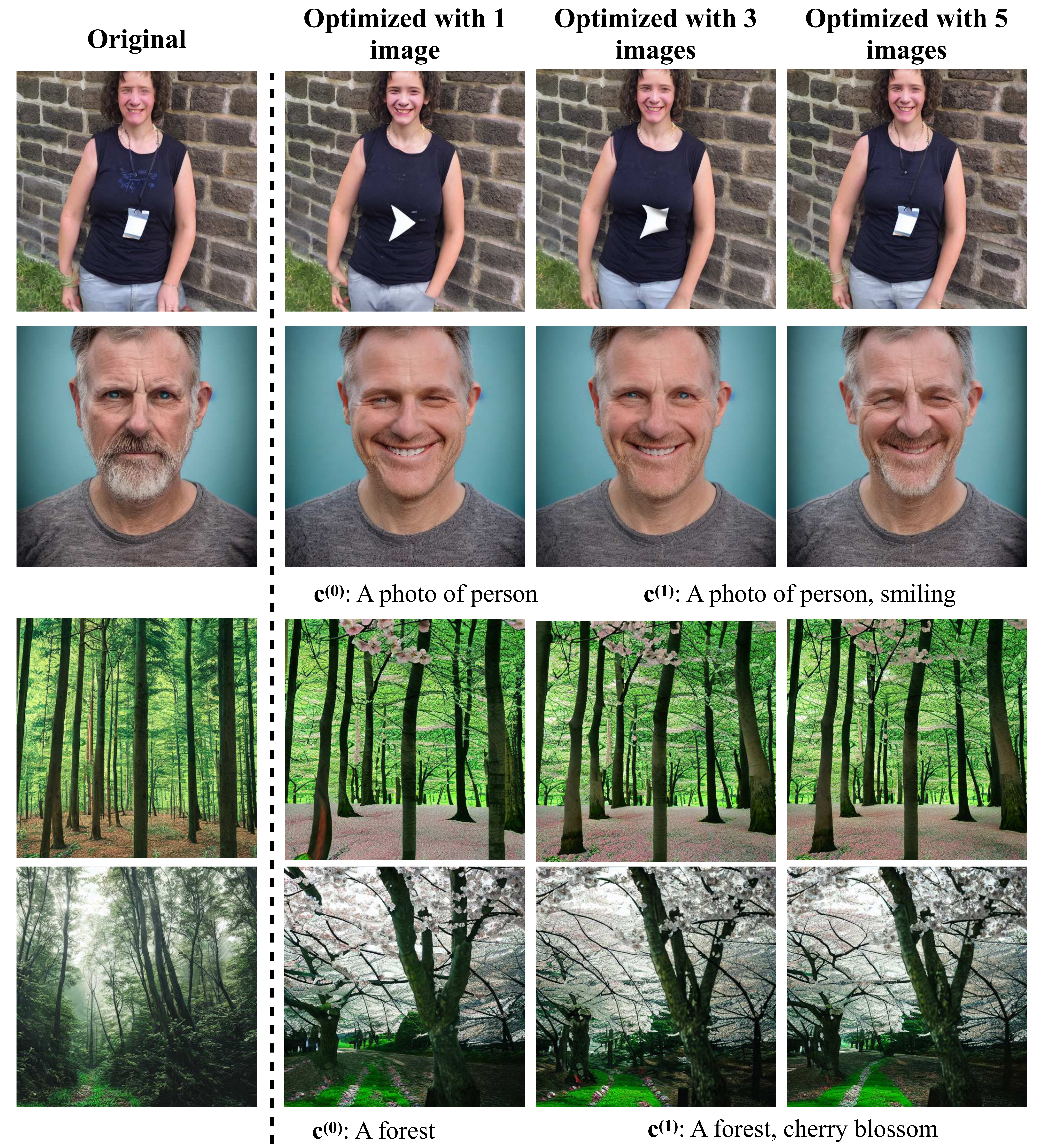}
\vspace*{-0.4em}
\caption{
\textbf{Results of using different numbers of images for optimization.}
The left column is the original unseen image, and the remaining columns demonstrate the results of applying the weights optimized on 1, 3, and 5 images respectively.
}
\vspace*{-0.1in}
\label{fig:supp-more-optimization}
\end{figure*}

\section{Effects of Varying Optimization Images}\label{optimization-image}
Finally, we investigate if our method is robust to the images used for optimization.
We examine whether a successful disentanglement depends on (1) the choice of a particular image used for optimization; and (2) number of images used for optimization.
Fig.~\ref{fig:supp-exp-H} illustrates the results for the first factor. For each attribute, we optimize \e{\lambda_{1:T}} on 3 different images separately. We then apply the resulting \e{\lambda_{1:T}} on the same unseen image and compare their transferring results.
As can be seen, different optimization images result in similar transferring results,
showing the robustness of our method to the choice of optimization image.
We further quantitatively measure the similarity of optimized combination weights by calculating their cosine similarity. As shown in Fig.~\ref{fig:6.4.1}, we visualize the similarity of \e{\lambda_{1:T}} between every pair of optimization images.
The high similarity score again demonstrates our method's robustness to the choice of optimization image.

The second factor is illustrated in Fig.~\ref{fig:supp-more-optimization}, where for each target attribute, we optimize \e{\lambda_{1:T}} on 1, 3, and 5 images respectively and show their transferring results on unseen images.
We observe that optimizing on more images leads to better disentanglement. For example, when optimizing on 5 images, the identity of the person is better preserved (\emph{e.g.,} the badge in the first row and the beard in the second row).
We therefore use 5 optimization images in Sec.~\ref{exp1}.

\end{alphasection}

\end{document}